\documentclass[fleqn,10pt]{wlscirep}
\usepackage[utf8]{inputenc}
\usepackage[T1]{fontenc}
\usepackage{xcolor}
\usepackage{todonotes}
\usepackage{comment}
\usepackage{pifont}
\usepackage{booktabs,multirow,makecell,colortbl,xcolor,pifont,array}
\usepackage{longtable}
\usepackage{array}
\usepackage{microtype}
\UseMicrotypeSet[protrusion]{basicmath}

\usepackage[table]{xcolor}
\definecolor{decrease}{HTML}{EE6C4D}
\definecolor{increase}{HTML}{00BBD6}
\colorlet{decrease}{decrease!30}
\colorlet{increase}{increase!15}

\usepackage[framemethod=tikz]{mdframed}
\usepackage{caption}
\usepackage[colorlinks=true]{hyperref}
\usepackage[nameinlink,capitalise]{cleveref}

\DeclareCaptionType{prompt}[Prompt][List of Prompts]

\newmdenv[
  backgroundcolor=white,
  linecolor=black!70,
  linewidth=0.6pt,
  roundcorner=6pt,
  innertopmargin=8pt,
  innerbottommargin=8pt,
  innerleftmargin=10pt,
  innerrightmargin=10pt
]{promptframe}

\usepackage{xcolor}
\usepackage{sectsty}
\usepackage{times}    
\usepackage{helvet}   
\usepackage{courier}

\definecolor{YaleBlue}{HTML}{00356B}

\usepackage{tcolorbox}
\tcbset{colback=gray!5!white, colframe=gray!40!black, fonttitle=\bfseries}

\usepackage{fontspec} 


\title{Rethinking Retrieval-Augmented Generation for Medicine: A Large-Scale, Systematic Expert Evaluation and\\Practical Insights}

\author[1]{Hyunjae~Kim}
\author[2]{Jiwoong~Sohn}
\author[3]{Aidan~Gilson}
\author[4]{Nicholas~Cochran-Caggiano}
\author[1]{Serina~Applebaum}
\author[5]{Heeju~Jin}
\author[5]{Seihee~Park}
\author[5]{Yujin~Park}
\author[5]{Jiyeong~Park}
\author[5]{Seoyoung~Choi}
\author[1,6]{Brittany~Alexandra~Herrera~Contreras}
\author[1]{Thomas~Huang}
\author[7]{Jaehoon~Yun}
\author[1,8]{Ethan~F.~Wei}
\author[1]{Roy~Jiang}
\author[1]{Leah~Colucci}
\author[1]{Eric~Lai}
\author[1]{Amisha~Dave}
\author[1]{Tuo~Guo}
\author[1]{Maxwell~B.~Singer}
\author[9]{Yonghoe~Koo}
\author[1]{Ron~A.~Adelman}
\author[10]{James~Zou}
\author[11]{Andrew~Taylor}
\author[12]{Arman~Cohan}
\author[1]{Hua~Xu}
\author[1,*]{Qingyu~Chen}

\affil[1]{Yale School of Medicine, Yale University, New Haven, CT, USA}
\affil[2]{Department of Biosystems Science and Engineering, ETH Zurich, Zurich, Switzerland}
\affil[3]{Massachusetts Eye and Ear, Harvard Medical School, Boston, MA, USA}
\affil[4]{Geisel School of Medicine at Dartmouth, Hanover, NH, USA}
\affil[5]{Seoul National University College of Medicine, Seoul, Republic of Korea}
\affil[6]{San Juan Bautista School of Medicine, Caguas, PR, USA}
\affil[7]{Hanyang University College of Medicine, Seoul, Republic of Korea}
\affil[8]{PA Leadership Charter School, West Chester, PA, USA}
\affil[9]{Asan Medical Center, University of Ulsan College of Medicine, Seoul, Republic of Korea}
\affil[10]{Stanford University School of Medicine, Stanford,
CA, USA}
\affil[11]{University of Virginia School of Medicine, Charlottesville, VA, USA}
\affil[12]{Yale School of Engineering \& Applied Science, Yale University, New Haven, CT, USA}
\affil[*]{qingyu.chen@yale.edu}

\begin{abstract}

Large language models (LLMs) are transforming the landscape of medicine, yet two fundamental challenges persist: keeping up with rapidly evolving medical knowledge and providing verifiable, evidence-grounded reasoning. 
Retrieval-augmented generation (RAG) has been widely adopted to address these limitations by supplementing model outputs with retrieved evidence. 
However, whether RAG reliably achieves these goals remains unclear. 
Here, we present the most comprehensive expert evaluation of RAG in medicine to date.
Eighteen medical experts contributed a total of 80,502 annotations, assessing 800 model outputs generated by GPT-4o and Llama-3.1-8B across 200 real-world patient and USMLE-style queries.
We systematically decomposed the RAG pipeline into three components---(i) \textit{evidence retrieval} (relevance of retrieved passages), (ii) \textit{evidence selection} (accuracy of evidence usage), and (iii) \textit{response generation} (factuality and completeness of outputs). Contrary to expectation, standard RAG often degraded performance: only 22\% of top-16 passages were relevant, evidence selection remained weak (precision 41--43\%, recall 27--49\%), and factuality and completeness dropped by up to 6\% and 5\%, respectively, compared with non-RAG variants. 
Retrieval and evidence selection remain key failure points for the model, contributing to the overall performance drop.
We further show that simple yet effective strategies, including evidence filtering and query reformulation, substantially mitigate these issues, improving performance on MedMCQA and MedXpertQA by up to 12\% and 8.2\%, respectively. These findings call for re-examining RAG's role in medicine and highlight the importance of stage-aware evaluation and deliberate system design for reliable medical LLM applications.

\end{abstract}

\begin{document}

\flushbottom
\maketitle
%
%
\thispagestyle{empty}

\newcommand{\draftcomment}[3]{%
  \todo[inline,color=#2!20]{\textbf{#1:}~#3}%
}

\newcommand{\hyunjae}[1]{\draftcomment{hyunjae}{teal}{#1}}

\section{Introduction}

Large language models (LLMs) have gained significant traction in medicine~\cite{tian2023opportunities,thirunavukarasu2023large,liu2025application}, with growing efforts focused on their deployment across a range of tasks, including medical question answering~\cite{lucas2024reasoning,yang2023integrating}, disease diagnosis~\cite{zhou2025large,liu2025generalist}, and treatment planning~\cite{kim2024llm,kim2024leme}. 
Yet, despite these advances, critical challenges persist in ensuring their safe and effective use in clinical settings. 
First, medical knowledge such as clinical guidelines and drug information is frequently revised as new evidence emerges~\cite{wu2025limitations, li2025reviewing}. 
To ensure clinical reliability, LLMs must continuously incorporate such updates to produce accurate and clinically relevant responses.  
A systematic evaluation of six leading LLMs---spanning the GPT~\cite{hurst2024gpt}, Gemini~\cite{team2024gemini}, and Llama families~\cite{grattafiori2024llama}---revealed marked limitations in addressing questions about newly approved drugs, with performance remaining consistently low~\cite{wu2025limitations}.
Second, in medicine, providing a decision or recommendation alone is rarely sufficient, and what often matters more is the ability to support that output with credible evidence~\cite{hager2024evaluation,li2024benchmarking}. 
Healthcare professionals routinely consult authoritative sources, particularly in uncertain or high-stakes scenarios where accountability and transparency are paramount~\cite{kell2024question,gallifant2025tripod}.
Current LLMs, however, lack explicit mechanisms for verification and typically generate responses without grounding in reliable references~\cite{kim2025medical,chen2025benchmarking,press2024citeme,wang2025medcite}. 
Earlier generations of LLMs frequently hallucinated in medical applications~\cite{kim2025medical,chen2025benchmarking}, and despite some progress, even the latest models still struggle to produce accurate citations or verifiable evidence~\cite{press2024citeme,wang2025medcite}.

Retrieval-augmented generation (RAG) has emerged as a promising paradigm to address these challenges~\cite{lewis2020retrieval,gupta2024comprehensive, amugongo2025retrieval}. 
RAG is designed to help models access up-to-date information and improve the factuality and trustworthiness of their responses, by incorporating external evidence at inference time~\cite{gupta2024comprehensive,jin2024genegpt}.
A standard RAG pipeline comprises three main stages~\cite{amugongo2025retrieval}.
First, domain-specific documents (e.g., clinical guidelines, biomedical publications, institutional protocols) are selected by practitioners, split into passages, and encoded into vector representations to build a searchable database. 
Second, given a user query, the system retrieves the top-$k$ most relevant passages (typically using cosine similarity between vector embeddings) and appends them to the input prompt. 
Third, the LLM integrates the retrieved passages with the query to generate the final response. 
This architecture offers notable advantages:
LLMs can access recent information without retraining, and users can provide domain-specific and authoritative knowledge sources from which the model retrieves evidence to support its outputs.
Owing to these strengths, RAG has attracted growing attention~\cite{fan2024survey,gargari2025enhancing} and now is widely adopted in medical applications, such as clinical question answering, trial screening, triage and diagnosis, and disease management~\cite{zakka2024almanac,unlu2024retrieval,kresevic2024optimization,wada2025retrieval,ke2025retrieval,gaber2025evaluating}.

However, despite its growing adoption, few studies have systematically examined how RAG performs in practice and whether it meaningfully addresses challenges related to factuality and verifiability. 
Most existing studies in medicine treat the RAG framework as a black box, applying it directly and evaluating only end-task performance, without analyzing intermediate steps such as retrieval quality or evidence usage—a gap also highlighted as a pressing need in recent reviews~\cite{yu2024evaluation,singh2025agentic}. 
Previous efforts have laid important groundwork but often remain constrained to pilot-scale studies (e.g., 30 queries in a specific medical specialty~\cite{gilson2024enhancing}) or examine only specific aspects such as citation relevance~\cite{wu2025automated}. 
Furthermore, recent studies have reported mixed results, with some suggesting that RAG may even reduce accuracy in downstream medical tasks~\cite{xiong2024benchmarking,sohn2025rationale}. 
These observations motivate a systematic investigation into how RAG interacts with LLMs in medical contexts. Several key aspects warrant in-depth examination: (1) whether the retriever retrieves evidence that is relevant to the query, since irrelevant documents may degrade response quality; (2) whether the LLM effectively identifies and uses the retrieved information; and (3) how retrieval-augmented responses compare to those generated without retrieval, particularly in medical decision-support scenarios.

\begin{figure}[t!]
\centering
\includegraphics[width=0.9\linewidth]{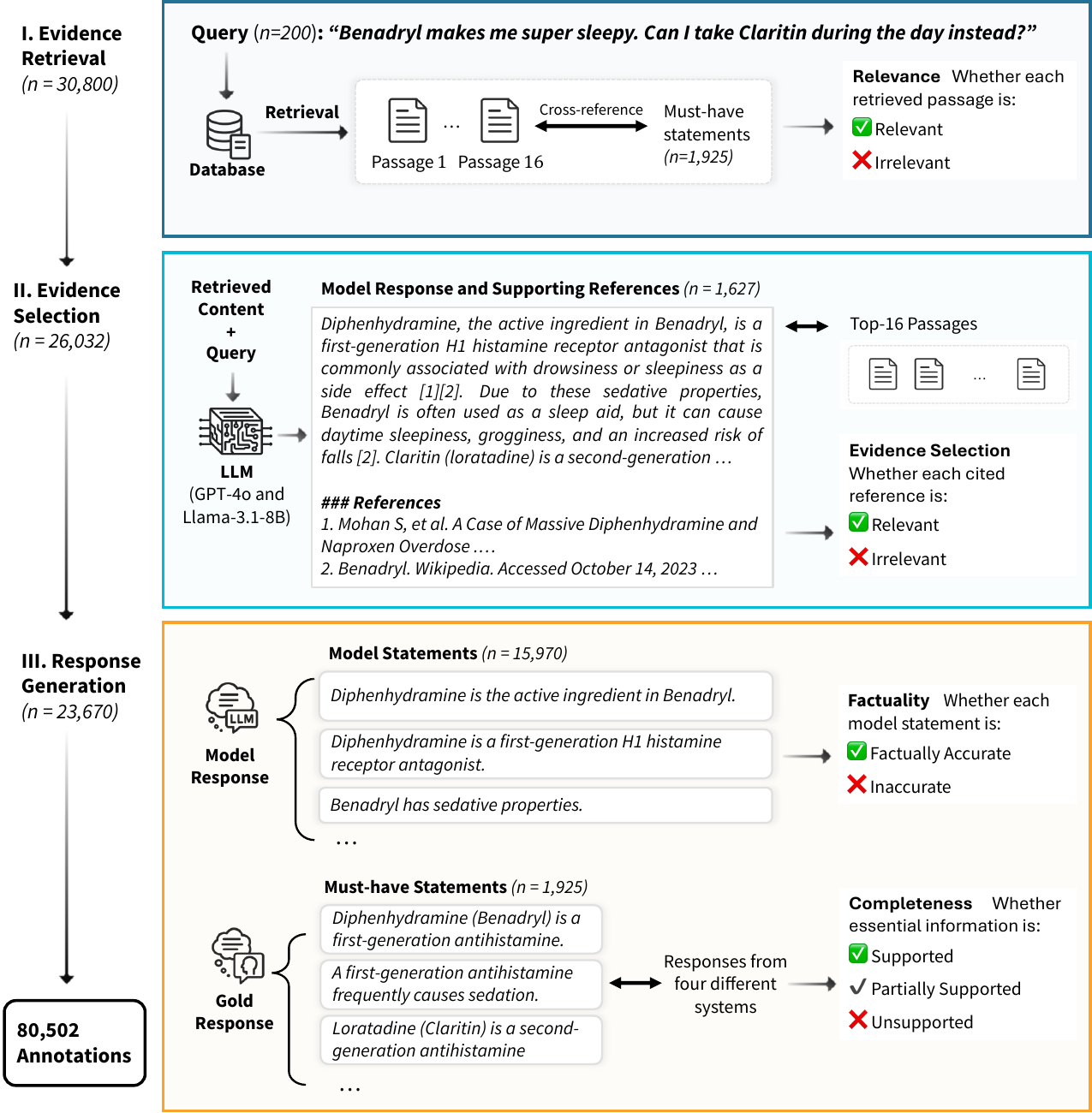}
\caption{
\textbf{Study design and evaluation framework.}
Our fine-grained framework decomposes the RAG pipeline into three components, evidence retrieval, evidence selection, and response generation, enabling systematic evaluation of each stage. 
First, retrieved passages are annotated for relevance (evidence retrieval; $n = $30,800). 
Next, model responses are evaluated to determine whether they are grounded in relevant evidence (evidence selection; $n = $26,032). 
Finally, responses are broken down into individual statements and assessed for correctness (factuality; $n = $15,970) and coverage of required information (completeness; $n = $7,700). 
In total, our framework comprises 80,502 expert annotations, enabling rigorous error attribution across the full RAG pipeline.
}
\label{figure:study_design}
\end{figure}

We conducted a large-scale, expert-driven evaluation to systematically assess the effectiveness of RAG in medicine. 18 medical professionals manually evaluated and contributed a total of 80,502 expert annotations across four model configurations---GPT-4o~\cite{hurst2024gpt} and Llama-3.1-8B~\cite{grattafiori2024llama}, each with and without RAG. The evaluation covered 200 medical queries, comprising 100 real free-text patient queries and {100 complex, context-rich scenarios modeled after the United States Medical Licensing Examination (USMLE).
As shown in Fig.~\ref{figure:study_design}, annotators evaluated RAG-LLMs across three stages:  
{(I) Evidence retrieval}: We assessed the relevance and coverage of the retrieved passages by determining whether they provided sufficient information to logically infer the key elements necessary for a correct answer.  
{(II) Evidence selection}: We evaluated whether LLMs effectively incorporated retrieved passages into their responses. 
Specifically, we measured the proportion of retrieved documents cited in the responses and how many of these were judged relevant in the evidence retrieval stage. 
{(III) Response generation}: We compared the final outputs of LLMs with and without RAG in head-to-head evaluations, focusing on factuality and completeness. 
Evaluations were performed at both the response level (i.e., the overall factuality and completeness of the full answer) and the statement level (i.e., the factuality and completeness of each individual statement).

Our findings reveal substantial limitations in the use of RAG for medical applications. Although RAG is widely assumed to improve response quality, it does not consistently enhance factuality or evidence usage, and in some cases, it may even diminish the overall quality of the generated responses.
{(I) Evidence retrieval}: The top-retrieved passages were often irrelevant and provided limited coverage of medical queries. 
On average, only 22\% of the top-16 passages were judged relevant, and the proportion was even lower for USMLE queries (15\%). 
In terms of coverage, the top-16 passages together contained information for only 33\% of must-have statements, dropping to 26\% for USMLE queries. 
{(II) Evidence selection}: Even when relevant passages were retrieved, LLMs often failed to incorporate them effectively. Both precision and recall for selecting relevant evidence remained low.
For GPT-4o, precision was 41\% and recall 49\%; for Llama-3.1, precision was 43\% and recall only 28\%. 
As a result, the already limited coverage of must-have statements from retrieval (33\%) dropped by nearly 10\% after evidence selection. 
At the same time, irrelevant passages were frequently included, with their frequency nearly twice that of relevant passages.
{(III) Response generation}: RAG may not improve LLM outputs. 
At both the response and statement levels, factuality and completeness were consistently lower for GPT-4o and Llama-3.1 with RAG than without. 
Response-level factuality dropped by up to 6\% for GPT-4o, while statement-level completeness declined by more than 5\% for Llama 3.1.
These degradations were more pronounced when the RAG system incorporated irrelevant passages or failed to retrieve relevant ones. For instance, Llama-3.1's factuality dropped by over 8\% when irrelevant passages were cited, compared to outputs grounded in relevant content. Similarly, completeness declined by over 6\% when relevant information was not retrieved at all.

We proposed two simple and practical strategies designed to directly mitigate the observed issues. 
First, evidence filtering removes irrelevant passages from the retrieved set, which is a necessary step given the high proportion of irrelevant content and the models' tendency to mistakenly incorporate it. 
Second, query reformulation rewrites the initial query to guide retrieval toward more relevant evidence, thereby improving the precision and coverage of the retrieval step. 
We evaluated each strategy independently and in combination across five medical question-answering benchmarks. In line with earlier findings, the RAG models did not consistently improve performance and in some cases reduced accuracy. 
In contrast, combining evidence filtering and query reformulation yielded substantial gains on more challenging datasets: Llama-3.1 improved by +12\% on MedMCQA and +8.2\% on MedXpertQA, while GPT-4o achieved gains of +3.4\% and +6.6\% on the same datasets.

Our study motivates a re-examination of RAG in the medical domain. 
Despite its promise, our study shows that RAG pipelines can introduce new sources of failure, including the retrieval of irrelevant information, failure to integrate relevant evidence, and reductions in both the factual accuracy and completeness of model outputs. 
By systematically evaluating each stage of the RAG process with a large number of expert annotations, we identify critical bottlenecks that have been largely overlooked in previous studies. 
Importantly, we demonstrate that targeted interventions, such as evidence filtering and query reformulation, substantially improve performance on challenging medical tasks. 
These findings suggest that the path forward lies not in applying RAG as a default solution but in rethinking both its system design and its evaluation.

\section{Results}

\subsection{Annotation Summary}
\label{section:annotation_summary}
We briefly summarize the annotation process (Fig.~\ref{figure:study_design}) and associated statistics below; full details, including stage-specific guidelines, annotator instructions, and interface design, are provided in the Methods section.

\subsubsection{Model Selection}
We conducted a comprehensive survey of current RAG practices in medicine, summarized in Supplementary Table~\ref{tab:rag_summary}, covering retrievers, LLMs, and knowledge sources. 
Our review includes 29 papers published between June 2023 and August 2025.
Among the retrievers, MedCPT~\cite{jin2023medcpt}, an open-source, domain-specialized retriever for biomedicine, was the most commonly used (28\%). 
Given its scalability and strong domain relevance, we selected MedCPT as the retriever for our pipeline.
Further discussion on alternative retrievers is provided in the Discussion section.
For LLMs, the majority of studies (over 65\%) relied on proprietary models from the GPT family (GPT-3.5, GPT-4, and GPT-4o), while a smaller subset employed open-source models, most notably variants of Llama.
In our study, we adopted both proprietary and open-source LLMs, GPT-4o and Llama-3.1-8B, respectively, to ensure a balanced evaluation across model types.
For the retrieval corpora, we included a broad range of widely used medical sources, including PubMed, Wikipedia, clinical guidelines, StatPearls, and medical textbooks.

\subsubsection{Data Curation}

We randomly sampled 100 patient queries from the K-QA dataset~\cite{manes2024k} and 100 USMLE-style questions from the MedBullets dataset~\cite{chen2025medbullets}. Both datasets provide gold-standard responses annotated by human experts.
In K-QA, each gold response is divided into ``must-have'' and ``nice-to-have'' statements, both verified by human annotators.
Note that these annotations were created independently of our study and originate from the original datasets.
The must-have statements capture the essential factual information required for a correct answer, whereas the nice-to-have statements provide supportive but non-essential details.
In contrast, the MedBullets dataset does not provide annotated statements. 
To address this, we prompted GPT-4o to extract statements from each gold response and classify them as must-have or nice-to-have, following prior work demonstrating the strong performance of LLMs in this task~\cite{wu2025automated}.
We used only the must-have statements for evaluating evidence retrieval and completeness scores, yielding a total of 1,925 statements across the two datasets.

For a total of 200 questions, we generated 800 responses using four model configurations: GPT-4o, GPT-4o with RAG, Llama-3.1, and Llama-3.1 with RAG. 
For the RAG-based models, the top-$k$ retrieved passages were provided as input context.
We set $k = 16$ to limit the annotation workload to a manageable level while still ensuring sufficient context for model reasoning.
Each response was accompanied by a list of supporting references, that is, source materials or citations the model cited as evidence for its answer.
For annotation, we exclusively used the 1,627 references produced by the RAG models. 
Each response was then segmented into individual statements using GPT-4o to enable statement-level factuality assessment. 
During this process, any inline citations (e.g., [1][2]) provided by the model were also separated along with each statement to enable later tracing of which reference supported which statement.
Non-factual claims that could not be objectively verified were excluded, resulting in a final total of 15,970 model-generated statements.

\subsubsection{Annotation Procedure and Statistics}
We conducted a multi-stage annotation process using a curated dataset comprising 200 queries, their top-16 retrieved passages, 1,925 must-have statements, 15,970 model-generated statements from 800 responses, and 1,627 references.
The paragraphs below describe how the full set of 80,502 annotations was derived by aggregating results from the three evaluation stages.

\paragraph{Evidence retrieval}
For each of the 200 queries, the retriever returned the top-16 passages.
Annotators then evaluated each passage by comparing it to the corresponding must-have statements, labeling it as relevant or irrelevant.
A passage was labeled as relevant if it fully (or partially) supported at least one of the must-have statements associated with the query; otherwise, it was labeled irrelevant.
This process yielded a total of 30,800 passage–statement pairs, computed as 1,925 must-have statements (aggregated across 200 queries) each evaluated against the top-16 retrieved passages for its corresponding query (1,925 x 16 = 30,800).

\paragraph{Evidence selection}
Annotators compared the 1,627 references generated by the two RAG models against each of the top-16 retrieved passages per query, identifying which passages corresponded to the sources actually cited by the model.
A single reference could be matched to multiple retrieved passages when appropriate, as multiple passages might originate from the same document or from different documents containing overlapping content.
This involved checking metadata (e.g., title, source name, URL) and reviewing the citation context in the model's response to confirm alignment between the cited reference and the retrieved passage.
For example, as shown in Fig.~\ref{figure:study_design}, if the model cited a PubMed article titled ``\textit{A Case of Massive Diphenhydramine and Naproxen Overdose},'' annotators assessed whether any retrieved passage matched that article based on both metadata and textual content.
Once the alignments were established, this provided a basis for evaluating whether each reference was grounded in a relevant or irrelevant source, using relevance labels from the previous stage.
This process produced a total of 26,032 passage–reference alignment annotations, calculated as 1,627 references evaluated against 16 retrieved passages per query (1,627 x 16 = 26,032).

\paragraph{Response generation} 
This stage was evaluated along two axes: factuality and completeness.
The factuality task assessed the correctness of individual statements generated by all four models, yielding a total of 15,970 statement-level annotations corresponding to all model-generated statements across queries.
The completeness task evaluates whether each model response adequately captures the essential content of the gold-standard response. 
For each of the 1,925 must-have statements, annotators judged whether the model's response fully supported, partially supported, or failed to support it, producing 7,700 response-statement annotations (1,925 statements x 4 models = 7,700).

\subsubsection{Inter-annotator Agreement}
Annotation reliability was assessed using Krippendorff's~$\alpha$ (nominal), with 10,000 bootstrap samples for confidence intervals.
Each sample in this analysis was independently labeled by two annotators.
For evidence retrieval, \textit{support} and \textit{partial support} labels were merged into a single category to address class imbalance during agreement estimation.
This yielded 1,280 annotated passage-statement pairs, with a resulting Krippendorff's~$\alpha$ of 0.757 (95\% CI: 0.638–0.852), indicating substantial agreement.
For evidence selection, a total of 416 reference-passage pairs were evaluated, resulting in a Krippendorff's~$\alpha$ of 0.926 (95\% CI: 0.906–0.944), indicating substantial inter-annotator consistency.
For the factuality criterion, 972 statements generated across the four model conditions were evaluated, yielding moderate inter-annotator agreement with a Krippendorff's~$\alpha$ of 0.512 (95\% CI: 0.372–0.636).
Although factuality judgments are grounded in objective evidence, they may still vary depending on each clinician's background knowledge and experience.
Importantly, each annotator assessed all model outputs for a given query, ensuring internal consistency within each case.
Thus, while the absolute agreement level was moderate, relative comparisons between models (e.g., RAG vs. non-RAG) remain reliable.
For the completeness criterion, 328 response–statement pairs were evaluated.
As with evidence retrieval, \textit{partial support} labels were merged with \textit{support} labels.
This yielded a Krippendorff's~$\alpha$ of 0.742 (95\% CI: 0.666–0.812), again indicating substantial inter-annotator agreement.

\subsection{Evidence Retrieval}

We employed three complementary metrics, all computed based on expert annotations of the top-16 retrieved passages per query. 
(1)~Precision@k measures the proportion of relevant passages among the top-k retrieved results. 
Here, a passage is considered relevant if it partially or fully supports at least one of the must-have statements associated with the query.
(2)~Miss@k reflects the proportion of queries for which no relevant passage was retrieved within the top-k. 
(3)~Coverage@k quantifies the proportion of must-have statements that were supported by at least one of the top-k retrieved passages.

\begin{figure}[t!]
\centering
\includegraphics[width=1.0\linewidth]{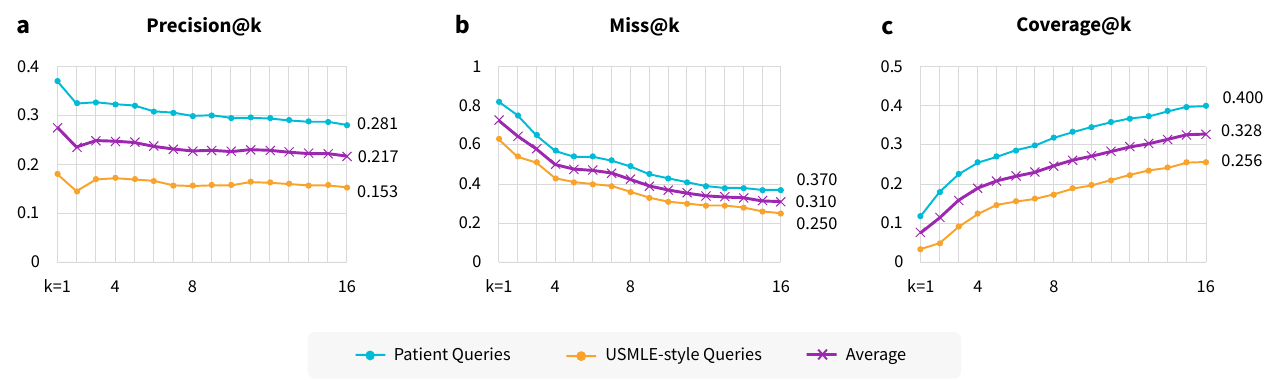}
\caption{
\textbf{Evidence retrieval performance across different evaluation metrics and query types.}
\textbf{a}, Precision@k: proportion of relevant passages among the top-k; higher is better.
\textbf{b}, Miss@k: proportion of queries with no relevant passage in the top-k; lower is better.
\textbf{c}, Coverage@k: proportion of must-have statements supported by the top-k; higher is better.
}
\label{figure:retrieval_results}
\end{figure}

Retrieval performance was markedly limited, with most retrieved passages failing to provide relevant support.
(1)~Precision@k remained low across all top-k values (Fig.~\ref{figure:retrieval_results}a). 
At ${k}=16$, precision was 0.217, indicating that only~\textasciitilde22\% of passages were identified as relevant. 
Performance was even lower for USMLE-style queries (0.153) compared to patient queries (0.281). 
This discrepancy reflects the dense contextual details in USMLE-style queries, such as medical history and symptom descriptions, which can hinder retrieval accuracy.
(2)~Miss@k revealed that a substantial fraction of queries failed to retrieve any relevant evidence, even at high top-k (Fig.~\ref{figure:retrieval_results}b).
Specifically, 31\% of queries had zero relevant passages among the top-16 retrieved results, highlighting the challenge of surfacing useful external information in complex clinical scenarios.
The failure rate was even higher for USMLE-style queries (37\%) than for patient queries (25\%). 
Notably, the issue was more pronounced at lower top-k, for instance, at top-1, 82\% of USMLE-style queries and 63\% of patient queries returned no relevant evidence.
(3)~Coverage@k of essential content was also limited (Fig.~\ref{figure:retrieval_results}c).
Here, essential information refers to must-have statements, which are manually annotated parts of the gold-standard answer required for a correct response.
The score was 0.328, indicating that only~\textasciitilde33\% of must-have statements were supported by the top-16 retrieved passages. 
While coverage gradually improved with larger top-k, a substantial portion of essential content remained unretrieved.
Once again, performance diverged by query type: USMLE-style queries achieved only 0.256 coverage, compared to 0.400 for patient queries.

\subsection{Evidence Selection}

We evaluated whether models effectively cited relevant information from the top-16 retrieved passages, which included a mix of relevant and irrelevant content.
On average, GPT-4o generated 4.9 references per query, while Llama-3.1 generated 4.5 (Fig.~\ref{figure:selection_results}a). 
Each reference was either retrieval-based---drawn directly from the top-16 passages---or self-generated using the model's internal knowledge.
The majority of GPT-4o’s references (89.8\%, 4.4 per query) were retrieval-based, compared to Llama-3.1 (62.2\%, 2.8 per query), indicating a higher proportion of self-generated references in the latter.
We computed micro-averaged precision and recall by comparing all retrieval-based references against expert relevance annotations. 
Self-generated references were excluded from this analysis, as annotations were available only for the retrieved passages.

Models struggled to selectively incorporate relevant evidence from the retrieved passages, as reflected in the low precision and recall scores across both models (Fig.~\ref{figure:selection_results}b).
GPT-4o achieved a precision of 0.412 and a recall of 0.486, while Llama-3.1 achieved a similar precision of 0.430 but a substantially lower recall of 0.275.
The low precision indicates that models frequently incorporated irrelevant content.
On average, GPT-4o cited 2.6 irrelevant passages per query out of 4.9 total references, while Llama-3.1 cited 1.6 out of 4.5 (Fig.~\ref{figure:selection_results}a).
This indicates that even high-performing models like GPT-4o often treat irrelevant content as valid evidence, reflecting limitations in their ability to distinguish useful information from misleading content.
In contrast, the low recall highlights the models' failure to effectively incorporate relevant information, even when it was readily available.
Based on the earlier precision@16 of 0.217, each query included approximately 3.5 relevant passages among the top-16 retrieved.
GPT-4o cited 1.8 of these on average, while Llama-3.1 cited just 1.2, indicating that both models failed to make full use of available evidence, with Llama-3.1 omitting nearly two-thirds of the relevant passages.
In both models, the number of relevant passages cited per query was lower than the number of irrelevant ones: 1.8 vs. 2.6 for GPT-4o and 1.2 vs. 1.6 for Llama-3.1.

\begin{figure}[t!]
\centering
\includegraphics[width=0.8\linewidth]{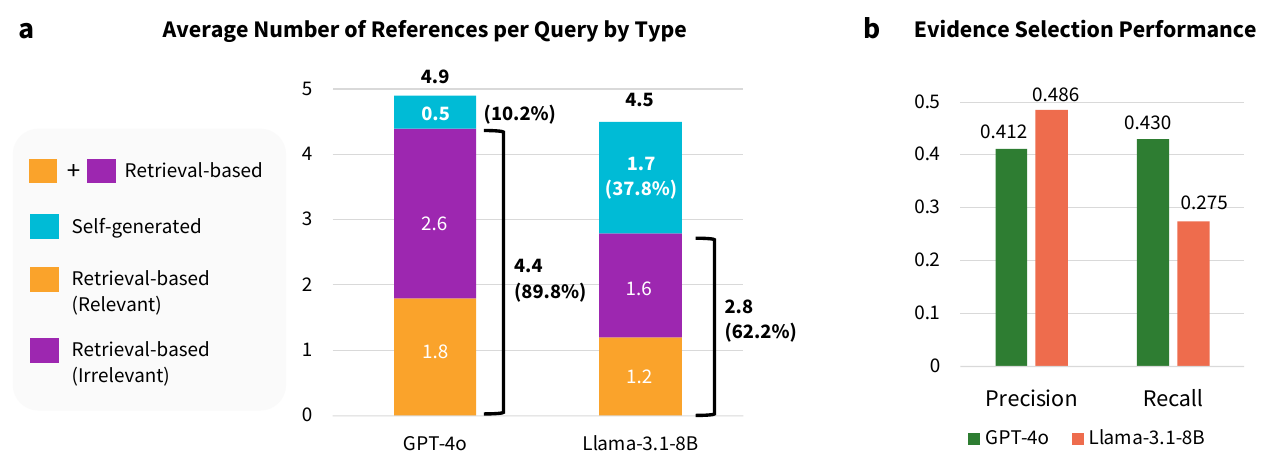}
\caption{
\textbf{Analysis of citation types and evidence selection performance.}
\textbf{a}, Average number of references per query, categorized by evidence source (retrieval-based vs. self-generated); retrieval-based references are further broken down by relevance.
Self-generated references refer to citations produced by the RAG model itself that do not appear among the retrieved passages. 
\textbf{b}, Precision and recall for identifying relevant evidence among retrieved passages. 
}
\label{figure:selection_results}
\end{figure}

\paragraph{Verifiability of self-generated references}
Beyond retrieved content, both models also produced references not found among the top-16 retrieved passages. 
These self-generated references made up 10.2\% of all citations in GPT-4o and 37.8\% in Llama-3.1 (Fig.~\ref{figure:selection_results}a). 
To verify their authenticity, we manually checked self-generated references from GPT-4o and Llama-3.1 to determine whether they could be located online using the generated citation details.
A small fraction (13.3\%) of GPT-4o's self-generated references were unverifiable, whereas a much larger proportion (77.2\%) of those from Llama-3.1 could not be confirmed, often containing fabricated metadata such as non-existent titles, authors, or publication sources. 
This discrepancy underscores a key limitation of small models: while they more frequently produce self-generated citations, they are also more prone to hallucinating plausible-sounding but non-existent references, raising concerns about the verifiability of their outputs.

\subsection{Response Generation}

We evaluated model outputs using two key criteria: factuality, which measures the correctness of the presented information, and completeness, which assesses whether essential information is fully conveyed.
All scores were manually determined through expert annotation.
For the RAG models, responses were generated using the top-16 retrieved passages.

\subsubsection{Factuality}
We evaluated factuality at two levels: response and statement.
At the response level, each answer was assigned a binary score---1 if the entire response was factually correct, and 0 if any part contained a factual error.
At the statement level, each individual statement within the response was evaluated independently as true or false, and we computed the proportion of factual statements for each query.
The response-level and statement-level scores were both averaged across all queries.

Factuality performance declined under the RAG setting for both models (Fig.~\ref{figure:overall_results}a).
At the response level, GPT-4o exhibited a 6.0\% drop in factuality, while Llama-3.1 showed a smaller decrease of 1.0\%.
At the statement level, both models experienced comparable declines: 1.6\% for GPT-4o and 1.9\% for Llama-3.1.
These results suggest that RAG negatively affects models across different scales and capabilities. 
Notably, GPT-4o showed a sharper decline at the response level, indicating that factual errors emerged more broadly across queries.
However, since GPT-4o started from a much higher baseline (68.0\%) compared to Llama-3.1 (39.5\%), the smaller drop observed in Llama-3.1 does not necessarily indicate greater robustness.

\begin{figure}[t!]
\centering
\includegraphics[width=\linewidth]{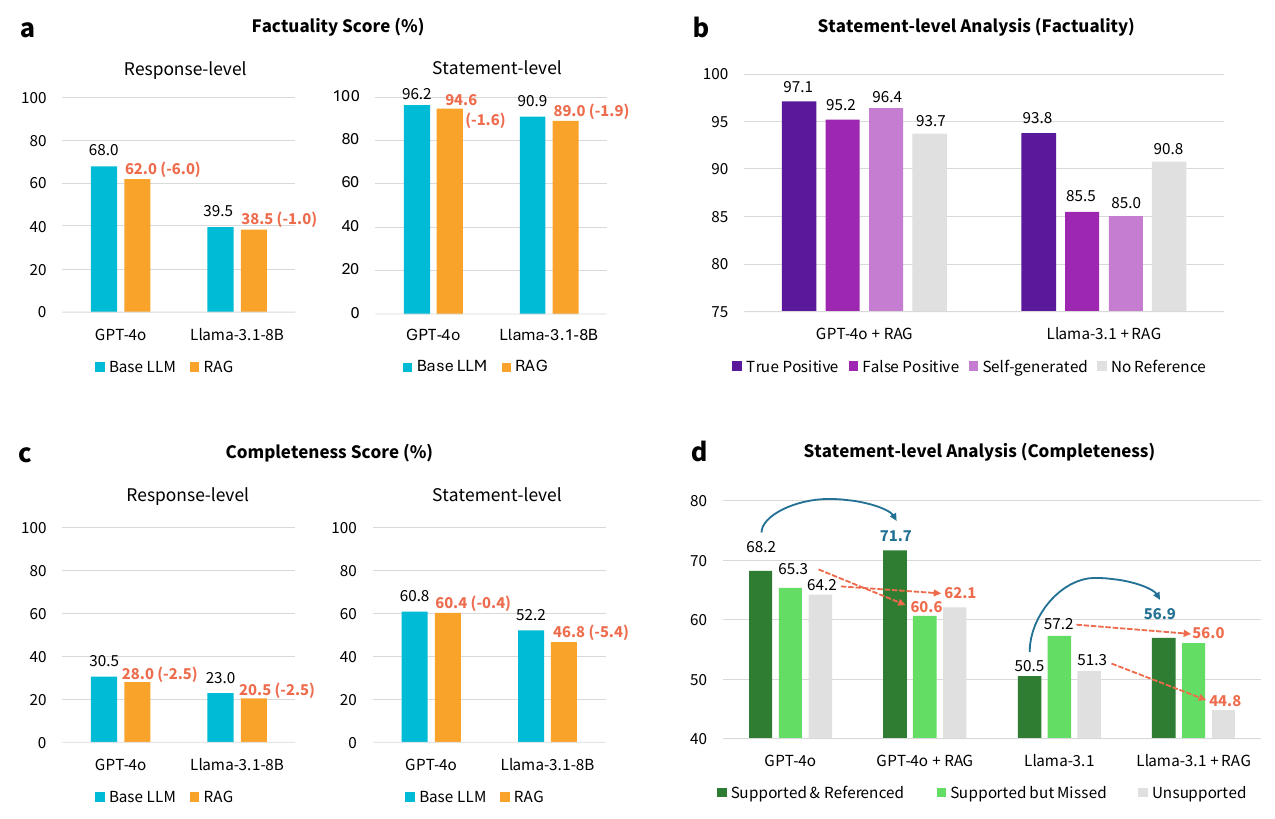}
\caption{
\textbf{Factuality and completeness of model responses.}
\textbf{a}, Average factuality scores at the response and statement levels across queries.
\textbf{b}, Statement-level factuality broken down by the type of evidence cited: relevant retrieved (True Positive), irrelevant retrieved (False Positive), self-generated, or none.
\textbf{c}, Average completeness scores at the response and statement levels, based on coverage of must-have statements.
\textbf{d}, Completeness for must-have statements, categorized by whether the supporting evidence was retrieved and cited (Supported \& Referenced), retrieved but not cited (Supported but Missed), or not retrieved at all (Unsupported).
}
\label{figure:overall_results}
\end{figure}

To better understand how evidence retrieval and selection affect model performance, we analyzed factuality at the statement level by grouping each statement based on the type of evidence cited by the model.
Specifically, for every statement produced by a RAG model, we identified whether it was supported by (1) a retrieved relevant passage (True Positive), (2) a retrieved but irrelevant passage (False Positive), (3) a self-generated source without retrieval grounding (Self-generated), or (4) no reference at all (No Reference).
As shown in Fig.~\ref{figure:study_design} (see II. Evidence Selection), references were explicitly marked within the model's response using numbered citation-style indicators (e.g., [1], [2]), each corresponding to a retrieved passage. 
By linking each statement to its cited references, and using our manual relevance annotations for the retrieved passages, we assigned every statement to one of the four evidence categories.

Factuality was highest when statements were grounded in relevant evidence, and declined when models relied on irrelevant or self-generated sources (Fig.~\ref{figure:overall_results}b).
When citing true positive passages, GPT-4o achieved a factuality of 97.1\%, while Llama-3.1 reached 93.8\%.
When citing false positives, GPT-4o maintained relatively high performance with 95.2\%, whereas Llama-3.1 dropped more substantially to 85.5\%.
This suggests that GPT-4o is more robust to noisy or off-topic evidence, often generating factually accurate content even when the cited material is not directly relevant. 
In contrast, Llama-3.1 appears more sensitive to the quality of retrieved content.
When models relied on self-generated evidence without any retrieved support, the lowest performance was observed. 
GPT-4o again maintained a high factuality score of 96.4\%, while Llama-3.1 fell to 85.0\%, indicating Llama-3.1's greater vulnerability to hallucination in the absence of external grounding.

\subsubsection{Completeness}
At the response level, each response was assigned a score of 1 if it successfully addressed all must-have statements defined in the gold-standard reference; otherwise, it received a score of 0.
At the statement level, each response was assessed for how many of the must-have statements it explicitly covered, and the proportion of addressed statements was calculated.
Both scores were averaged across all queries.

Completeness performance declined under the RAG setting for both models (Fig.~\ref{figure:overall_results}c).
At the response level, GPT-4o and Llama-3.1 each showed a 2.5\% decrease.
At the statement level, GPT-4o exhibited a minimal decline of 0.4\%, while Llama-3.1 experienced a more pronounced drop of 5.4\%.
These results suggest that incorporating retrieved evidence does not guarantee improved coverage of essential content and may even hinder completeness, particularly in smaller or less capable models.

We further analyzed model performance at the statement level by categorizing must-have statements according to the quality of the retrieved evidence and whether it was cited in the response.
Specifically, each must-have statement was grouped into one of three categories:
(1) Supported and Referenced: The statement was supported by one of the top-16 retrieved passages and explicitly referenced in the model's response.
(2) Supported but Missed: The statement was supported by a retrieved passage but was not cited by the model.
(3) Unsupported: The statement was not supported by any of the retrieved passages.

Completeness was highest when models correctly referenced relevant evidence, and declined when such evidence was either missed or not retrieved. (Fig.~\ref{figure:overall_results}d).
In the Supported and Referenced category, both models improved under the RAG setting: GPT-4o increased from 68.2\% to 71.7\%, and Llama-3.1 from 50.5\% to 56.9\%.
However, when models failed to cite available supporting evidence (Supported but Missed), GPT-4o's score decreased from 65.3\% to 60.6\%, and Llama-3.1 also experienced a slight decrease from 57.2\% to 56.0\%.
These results suggest that access to high-quality evidence alone is insufficient, and effective evidence selection remains critical for achieving high completeness.
In the Unsupported category, completeness declined across both models with RAG: GPT-4o fell from 64.2\% to 62.1\%, and Llama-3.1 from 51.3\% to 44.8\%.
This indicates that even with retrieval augmentation, models may struggle to compensate when key content is absent from the retrieved documents, especially in the case of Llama-3.1, which showed the steepest drop.

\paragraph{Answer accuracy} 
We evaluated answer accuracy on the full dataset from which the USMLE-style queries were originally sampled.
In line with our earlier findings on factuality and completeness, both GPT-4o and Llama-3.1 showed clear performance degradation after applying standard RAG at $\text{top-}k=16$, with accuracy dropping from 71.1\% to 70.5\% and from 49.4\% to 45.8\%, respectively.
These results suggest that the declines in factuality and completeness are not isolated errors, but reflect broader impairments in overall response quality.

\subsection{Enhanced RAG Pipeline}
\begin{figure}[t!]
\centering
\includegraphics[width=\linewidth]{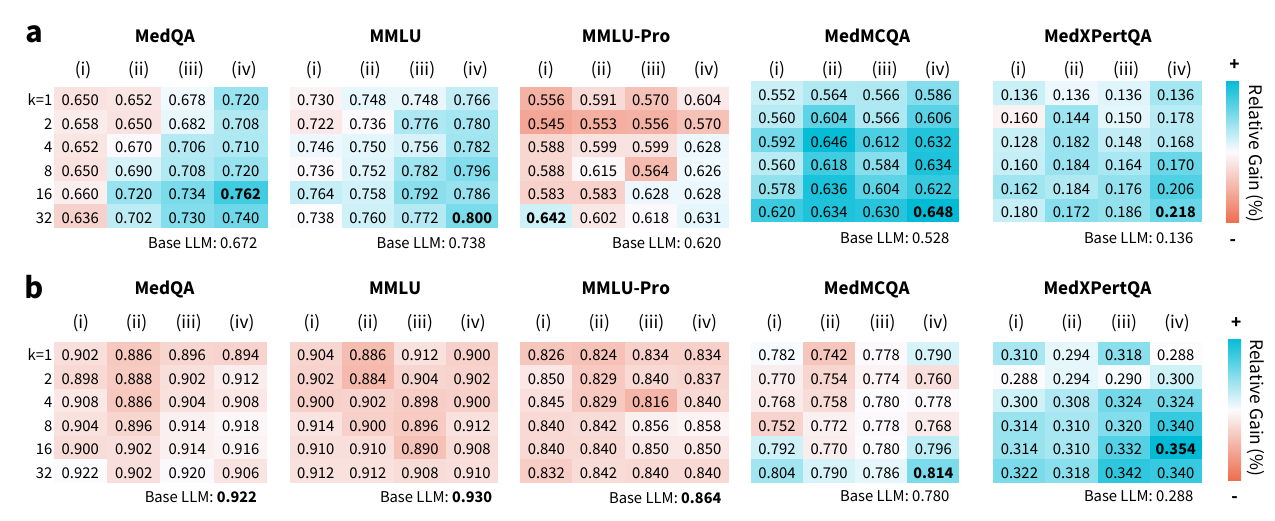}
\caption{
\textbf{Performance of RAG variants and non-RAG baselines across five QA datasets}.
MedQA~\cite{jin2021disease}, MMLU~\cite{hendrycks2021measuring}, MMLU-Pro~\cite{wang2024mmlu}, MedMCQA~\cite{pal2022medmcqa}, and MedXPertQA~\cite{zuo2025medxpertqa} were used.
\textbf{a}, Results using Llama-3.1-8B as the base model.
\textbf{b}, Results using GPT-4o as the base model.
Cell colors indicate the magnitude and direction of accuracy changes relative to the base LLM: sky-blue denotes performance gains, while red indicates performance drops.
Four RAG configurations are compared: (i) standard RAG (retrieval-only), (ii) retrieval + evidence filtering, (iii) retrieval + query reformulation, and (iv) retrieval + both evidence filtering and query reformulation, evaluated at different top-k settings ($\text{k}=1, 2, 4, 8, 16, 32$).
Each cell displays the accuracy of the model, while the color represents the relative gain compared to the base LLM. 
Positive gains (blue) indicate improved performance over the base LLM, while negative values (red) indicate performance drops. 
Base LLM accuracies are shown beneath each block for reference.
The best scores are highlighted in bold.
}
\label{figure:qa_results}
\end{figure}

We evaluated the impact of retrieval augmentation using five held-out QA datasets on GPT-4o and Llama-3.1. 
Four configurations were tested: (i) standard RAG, which retrieves and appends passages to the input; (ii) RAG with evidence filtering, which removes irrelevant passages to suppress retrieval noise; (iii) RAG with query reformulation, which addresses low coverage in the initial retrieval by using the model’s first response as a rewritten query; and (iv) a combined pipeline integrating both filtering and reformulation.
The filtering module was fine-tuned on our expert-annotated relevance dataset using Llama-3.1-8B as the backbone, while the reformulation component adopted a rationale-driven querying strategy to generate more targeted follow-up queries (see Methods for details).
Accuracy was automatically computed, and confidence intervals were estimated via 10,000 bootstrap replicates sampled with replacement at the same size as each test set.
Detailed results are provided in the Supplementary Table~\ref{table:qa_result}.

Standard RAG produced inconsistent effects across datasets and retrieval depths (Fig.~\ref{figure:qa_results}a,~i).
On Llama-3.1, standard RAG led to accuracy gains in some cases (e.g., MedMCQA) but performance drops in others (e.g., MedQA). 
Even within the same dataset, outcomes varied by top-k: for example, MMLU-Pro showed degradation from $k=1$ to $k=16$ and improvement only at $k=32$. 
This instability highlights the sensitivity of standard RAG to retrieval noise and passage overload.
Additionally, standard RAG on GPT-4o also exhibited inconsistent effects, though performance degradation was even more pronounced (Fig.~\ref{figure:qa_results}b,~i).
On datasets where GPT-4o already achieved strong baseline accuracy, such as MedQA (0.922), MMLU (0.930), and MMLU-Pro (0.864), standard RAG led to a decline in performance across all top-k settings.
This suggests that for high-performing LLMs, the additional context can introduce more distraction than utility.
As with Llama-3.1, MedXPertQA was a notable exception, where standard RAG yielded some gains at higher top-k values.

Combining evidence filtering and query reformulation yielded consistent and significant improvements across all benchmarks (Fig.~\ref{figure:qa_results}a,~ii-iv).
Individually, both filtering and reformulation led to performance improvements in some settings, but failed to produce consistent gains across datasets. 
When integrated, however, the two components complemented each other---filtering reduced noise from irrelevant passages, while reformulation compensated for low coverage in retrieved evidence---resulting in stable and monotonic gains from $k=4$ to $k=32$. 
The combined pipeline outperformed the non-RAG baseline by up to +9.0\% (MedQA), +6.2\% (MMLU), +1.1\% (MMLU-Pro), +12.0\% (MedMCQA), and +8.2\% (MedXpertQA), with all improvements statistically significant (McNemar exact test, $p < 0.001$ for most and $p = 0.0022$ for MMLU) except MMLU-Pro.
For GPT-4o, the combined pipeline led to notable gains on the more challenging datasets, MedMCQA (+3.4\%) and MedXpertQA (+6.6\%) ($p = 0.013$ and $0.027$), whereas no improvement was observed on MedQA, MMLU, and MMLU-Pro, where where baseline accuracy was already high and left limited room for further gains (Fig.~\ref{figure:qa_results}b,~ii–iv).

\section{Discussion}

This study presents the first large-scale, fine-grained evaluation of widely used medical RAG frameworks to elucidate their behavior throughout the end-to-end pipeline.
We surveyed the literature to understand which RAG configurations are most commonly used in medical applications. Based on these representative frameworks, we evaluated two LLMs (GPT-4o and Llama-3.1-8B) on 200 queries (real patient inquiries and USMLE-style questions) across evidence retrieval, evidence selection, and response generation. 
Eighteen medical experts contributed over 80,000 expert annotations.
Such an extensive, expert-annotated analysis has not been previously achieved in the medical RAG literature.
Whereas prior efforts, such as the MIRAGE benchmark~\cite{xiong2024benchmarking}, focused primarily on answer accuracy across large question sets, these efforts offered limited insight into how and where failures emerge within the RAG pipeline. 
In contrast, our fully manual, component-level evaluation dissects the end-to-end process and highlights system weaknesses, directly addressing recent calls for more nuanced and clinically relevant RAG assessment~\cite{amugongo2025retrieval}.

First and importantly, our evaluation shows that standard RAG may not improve and can even degrade LLM performance on medical tasks. 
In head-to-head comparisons with their non-RAG counterparts, GPT-4o and Llama-3.1-8B both exhibited modest drops in factuality and more pronounced declines in completeness, with the latter decreasing by over five percentage points at the statement level. 
This pattern held across both query types:
for patient queries, GPT-4o's factuality and completeness dropped by approximately 2\% and 8\%, respectively;
for USMLE-style queries, the decreases were 1\% and 3\%.
While many prior studies have highlighted RAG's ability to improve factuality and mitigate hallucinations in medical applications~\cite{manes2024k,zakka2024almanac,miao2024integrating,luo2024development}, others--particularly in the general domain--have reported that irrelevant or noisy retrieved content can distract models or sustain hallucinations~\cite{ding2024retrieve,niu2024ragtruth,sun2025redeep}. 
This aligns with our findings: when the model incorporated irrelevant rather than relevant passages (i.e., the ``False Positive'' category in Fig.~\ref{figure:overall_results}), statement-level factuality scores declined, with the effect being especially pronounced for the smaller model, Llama-3.1, which exhibited a drop of over 8\%.
Completeness is another critical dimension where retrieval is generally expected to fill knowledge gaps and broaden the scope of model responses. 
However, prior evidence remains mixed. 
In long-form clinical QA tasks, one study reported improved completeness over non-RAG LLMs based on manual evaluation~\cite{zakka2024almanac}, whereas others observed declines--for example, a drop from 3.47 to 3.27 on a five-point scale in one study~\cite{gilson2024enhancing}, and a roughly 5\% decrease reported in another~\cite{manes2024k}.
Our results indicate a consistent shortfall: when no supporting passage was retrieved for a must-have statement (i.e., Unsupported), models often overlooked key information they had previously included without retrieval, leading to substantial performance degradation, with over 6\% for Llama-3.1. 
Interestingly, when a relevant passage was retrieved but not cited by the model (i.e., Supported but Missed), completeness still dropped by 1-5\%, suggesting that the model may have been distracted by other retrieved passages and failed to integrate critical evidence.

Through our stage-wise analyses, we highlight two primary bottlenecks underlying these declines. First, evidence retrieval is often inadequate: only 22\% of the top-16 passages were judged relevant, and 31\% of queries lacked any relevant passage at all. 
Prior studies have largely examined how irrelevant or conflicting retrieved content influences end-task performance~\cite{ding2024retrieve,niu2024ragtruth,cuconasu2024power,yoran2024making,wu2024easily,sun2025redeep,amiraz-etal-2025-distracting}; however, the retrieval accuracy for medical queries itself has rarely been evaluated in a standardized, expert-grounded manner. 
Our manual evaluation addressed this gap, showing that current retrievers frequently mishandled clinical questions. 
Moreover, the retrieved content covered only 33\% of must-have statements overall and just 26\% for USMLE-style queries. 
This indicates that even with RAG, LLMs often receive limited knowledge support and must rely primarily on their internal knowledge to address medical queries, highlighting the persistent challenge of providing sufficient context~\cite{joren2025sufficient,xie2025rag}.
Importantly, our notion of relevance differs from conventional definitions in information retrieval systems~\cite{wei2006lda,robertson2009probabilistic,lavrenko2017relevance,karpukhin2020dense,khattab2020colbert,zhao2024dense}.
Rather than requiring an explicit answer span, lexical overlap, or topical relatedness, we adopt a coverage-based criterion that evaluates whether a passage supplies the necessary medical information to support key statements.
This yields a more clinically grounded and functionally meaningful view of retrieval quality.

The second bottleneck is evidence selection, where LLMs struggle to identify and integrate relevant information even when it is retrieved. 
Both GPT-4o and Llama-3.1 struggled in selecting relevant evidence, achieving only 41-43\% precision and 27-49\% recall. 
GPT-4o tended to cite a larger number of retrieved passages (90\% of citations vs. 62\% for Llama-3.1), indicating a stronger inclination to incorporate retrieved content into its responses. 
This behavior presents a potential advantage: when provided with high-quality, relevant evidence, GPT-4o appears better equipped to integrate it effectively. 
However, this responsiveness comes at a cost—its limited evidence selection precision leads to a greater number of irrelevant references, averaging 2.6 per query compared to 1.6 for Llama-3.1.
In contrast, Llama-3.1 cited fewer retrieved passages and more frequently missed relevant content, selecting only 34\% of the relevant evidence available. 
It also relied more heavily on self-generated references (37.8\% of all citations), the majority of which (77.2\%) could not be verified and often included fabricated titles, authors, or publication sources.
These findings highlight complementary failure modes between model scales: while larger models like GPT-4o risk over-incorporating retrieved information without sufficient discrimination, smaller models like Llama-3.1 may under-utilize retrieved evidence and compensate by hallucinating references.
Our evidence selection analysis is closely related to the actively growing body of research on evidence attribution and citation generation, both of which aim to evaluate and improve the verifiability of LLM responses~\cite{rashkin2023measuring,gao2023enabling,liu2023evaluating,yue2023automatic,huang2024training,malaviya2024expertqa,asai2024openscholar,wu2025automated,zhang-etal-2025-longcite,wang-etal-2025-medcite}. 
Consistent with these studies, our findings confirm that current LLMs still exhibit notable limitations in reliably using available evidence and grounding their responses appropriately.
Additionally, we positioned evidence selection as a connecting layer within the medical RAG pipeline, enabling us to trace how retrieved content contributes to model responses at the statement level.
This represents a novel analytical perspective in the context of medical RAG, bridging retrieval and generation more explicitly than in previous studies.

To better understand why RAG underperformed compared to its non-RAG counterparts, we conducted a qualitative error analysis focusing on cases where RAG responses demonstrated lower statement-level factuality.
We observed that RAG often introduced errors by anchoring responses to misleading numerical references in the retrieved passages. 
For instance, in one example, a prolactin level of 28 ng/mL was initially judged normal by GPT-4o, but the RAG model flagged it as abnormal. 
This change appears to stem from retrieved information that defined the upper limit as 25 ng/mL; although this information was contextually misaligned with the original question, the model incorrectly adopted the reference range and judged 28 ng/mL as abnormal.
A similar issue emerged in response to the question, ``What is the safest amount of Advil to take at one time?'' 
The GPT-4o-based RAG model responded that 400 mg is the safest dose--an overstatement likely driven by the fact that multiple retrieved studies consistently referenced 400 mg as the standard single adult dose in bioequivalence and safety evaluations, though these studies were situated in different clinical or experimental contexts.
Another recurrent failure mode involved lexical ambiguity or synonym mismatches in retrieval. 
When asked ``Is poison ivy contagious between people?'', the retriever surfaced content about the character Poison Ivy from DC Comics and the 1992 film of the same name rather than about the allergenic plant. 
Consequently, Llama-3.1 generated a response describing the fictional character, entirely missing the medical intent of the question. 
This error illustrates both the retriever's failure to disambiguate terms and the model's inability to recover from such retrieval mistakes.
We also found that RAG models were more prone to misinterpreting user intent, especially when retrieved content introduced a competing frame. 
In response to the question ``How long does this year's flu usually last?'', both non-RAG models correctly addressed symptom duration at the individual level. 
However, once RAG was introduced, the models instead focused on the epidemiological length of flu seasons, which is an error likely caused by retrieved passages discussing historical influenza trends and population-level data. 
These examples collectively demonstrate that errors originating in the retrieval stage act as noise, impairing the accuracy and reliability of the model's generated responses.

Beyond evaluation, we proposed augmenting the standard RAG pipeline with two lightweight and model-agnostic components: evidence filtering and query reformulation. 
Unlike approaches that require retraining either the retriever or the LLM~\cite{asai2024self,jeong2024improving,jin2025search}, our methods are readily applicable to any retrieval systems and LLMs.
For evidence filtering, we first evaluated zero-shot performance using our 3,200 expert-annotated samples spanning 200 queries and their corresponding top-16 retrieved passages. 
Zero-shot filtering proved insufficient, yielding limited performance for both Llama-3.1 (precision = 0.483, recall = 0.566, F1 = 0.521) and GPT-4o (precision = 0.697, recall = 0.324, F1 = 0.442). 
After fine-tuning Llama-3.1 with five-fold cross-validation, performance improved substantially (precision = 0.592, recall = 0.657, F1 = 0.623).
These results highlight that filtering is effective but cannot be reliably achieved in a zero-shot setting. 
Supervised signal remains important, underscoring the value of our large-scale expert annotations as a reusable training resource for future modules.
For query reformulation, we drew inspiration from prior work showing that models capable of articulating explicit rationales tend to generate more focused and contextually appropriate queries~\cite{wang2023query2doc,jagerman2023query,sohn2025rationale}. 
Building on this rationale-guided formulation approach, we prompted the model to first produce an intermediate response to the initial query and then used that response itself as a reformulated query.
To assess the effect of reformulation, we compared retrieval outcomes between initial and reformulated queries using the MedQA dataset. For each query, we retrieved 256 passages and applied our fine-tuned evidence filtering model to assess relevance. 
On average, reformulated queries yielded substantially more passages deemed relevant than the original queries (32\% vs. 13\%), indicating that reformulation enhanced both the precision and contextual alignment of the retrieval stage.

Building on the same evaluation framework, we also explored whether stronger retrievers could further improve performance.
We evaluated two representative approaches: BM25~\cite{robertson2009probabilistic}, a classical lexical baseline widely used for its simplicity, and Qwen3-Embedding-0.6B~\cite{zhang2025qwen3}, an open-source LLM-based embedding model that outperforms commercial API-based models (\textit{text-embedding-ada-002}, \textit{text-embedding-3-large}) on the MTEB leaderboard~\cite{muennighoff2023mteb}~(\url{https://huggingface.co/spaces/mteb/leaderboard}).
As shown in Supplementary Table~\ref{table:qa_result}, no single retriever consistently outperformed others across the five medical QA datasets.
Performance varied by dataset, and even the most advanced model, Qwen3-Embedding-0.6B, did not achieve the highest accuracy overall.
These results indicate that retriever replacement alone cannot overcome the intrinsic limitations of the RAG architecture.
In contrast, our proposed modules (evidence filtering and query reformulation) yielded more robust and generalizable improvements across both BM25 and Qwen3 retrievers, consistent with our MedCPT-based findings.

Looking forward, we recommend that future RAG development in medicine move beyond end-task performance metrics and adopt a stage-wise perspective that explicitly examines retrieval, selection, and generation behaviors. 
Since RAG pipelines do not consistently lead to performance gains across tasks, a safer and more reliable approach is to incorporate lightweight modules such as evidence filtering and query reformulation, which can offer practical improvements without requiring retriever or model retraining.
Further research should explore how these interventions interact with different retrievers, search databases, medical subdomains, and LLM architectures, as well as how they influence factual grounding and reasoning consistency. 
We also observed that the benefits of RAG augmentation tend to diminish when the underlying model already demonstrates strong performance. 
In such cases, engineers may need to determine whether RAG should be applied using a small validation set. 
In the longer term, an adaptive architecture that selectively invokes RAG only when the model's internal knowledge appears insufficient could offer a promising and efficient solution.
Finally, establishing and sharing standardized expert-annotated datasets, such as those introduced in this study, will be essential for advancing medical RAG. While expert-driven evaluation remains critical, it should be complemented by more scalable strategies, including community-based review~\cite{chiang2024chatbot} and automated assessment~\cite{ru2024ragchecker}. These efforts will support more reliable and sustainable evaluation across diverse medical settings.

\section{Methods}

\subsection{Annotation}

\subsubsection{Data Curation}
\paragraph{Patient queries}
We used a sample of 100 queries from the K-QA dataset~\cite{manes2024k}
This dataset originally comprises 201 free-form clinical questions sourced from real-world patient–physician conversations on an online consultation platform.
The queries, submitted by patients, span a broad range of topics, including ear, nose, and throat (ENT), dermatology, mental and emotional health, and vision and eye care.
Each query is paired with a long-form, natural language response written by a physician.
In addition, the dataset includes manual annotations identifying must-have and nice-to-have statements within the physician responses.

\paragraph{USMLE-stytle queries}
We used a sample of 100 multiple-choice questions from the MedBullets dataset~\cite{chen2025medbullets}, which is curated for USMLE Step 2 and Step 3 preparation.
Unlike other datasets that provide only questions and answers~\cite{jin2021disease,hendrycks2021measuring}, MedBullets includes detailed, expert-authored explanations, offering richer context and rationale.

\paragraph{Statement extraction}
As the MedBullets dataset does not include statement-level annotations, we curated must-have statements by using GPT-4o to segment each explanation into individual statements.
We employed a few-shot prompting strategy, using a mixture of general- and medical-domain examples adapted from prior work~\cite{min2023factscore,manes2024k}.
The full prompt with examples is shown in Supplementary Table~\ref{tab:prompt_stat_extraction}.
The same procedure was also applied to segment model responses into individual statements for downstream evaluation.

\paragraph{Must-have statement identification}
The full set of extracted statements, along with the question and gold answer, was then passed to GPT-4o using a few-shot prompt to identify which statements were essential for answering the question, guiding the model to select only the most critical information.
The model produced a list of binary labels indicating whether each statement was considered must-have.
The full prompt with examples is shown in Supplementary Table~\ref{tab:prompt_must_have}.

\paragraph{Response generation}
We generated model responses using both RAG and non-RAG variants. 
We used identical prompts to generate responses from both GPT-4o and Llama-3.1-8B. 
However, different sampling parameters were applied to accommodate each model's behavior. 
Specifically, GPT-4o was run with \texttt{temperature = 0.8}, while Llama-3.1-8B used \texttt{temperature = 1.0} and \texttt{top\_p = 0.9}.
We intentionally avoided greedy decoding, as deterministic outputs often failed to comply with formatting requirements, particularly for citation generation. 
Instead, we introduced controlled randomness and, if necessary, performed multiple inference attempts to obtain outputs that conformed to the expected structure.
The full prompt used is shown in Supplementary Table~\ref{tab:prompt_response}.

\paragraph{Model statement filtering}
Some model statements could not be reliably assessed for factuality, as they lacked verifiable factual content.
For instance, they simply rephrased the question, provided generic commentary, or included procedural language.
To exclude such statements, we applied a GPT-4o classifier prompt to identify and filter out non-distinctive content.
A statement was considered distinctive only if it introduced new clinical reasoning, factual knowledge, or a definitive judgment not already stated or implied in the question.
The prompt used for this classification is shown in Supplementary Table~\ref{tab:prompt_model_stat_filtering}.

\paragraph{Statement-citation alignment}
We aligned inline citations with the individual statements extracted from each response.
Any inline citations were preserved alongside their corresponding statements, ensuring that each statement remained linked to its supporting reference (or none, if no citation was originally provided).
This process enabled us to later trace which reference was intended to support each claim.
The alignment was performed using GPT-4o, with the prompt details provided in Supplementary Table~\ref{tab:prompt_alignment}.

\subsubsection{Annotation Procedure}

\paragraph{Overview}
All annotations were conducted using a custom interface built on Label Studio (\url{https://labelstud.io/}), designed to provide annotators with a streamlined and user-friendly labeling workflow.
Annotators were given written annotation guidelines describing each task, along with example cases, and we held synchronous meetings prior to the annotation phase to clarify task definitions and resolve ambiguities.
A total of 18 annotators participated in this study. 
Of these, nine were assigned to the evidence retrieval and evidence selection tasks, and the remaining nine were assigned to the response generation tasks (factuality and completeness). 
All annotators involved in the response generation tasks were either medical residents or clinical fellows, ensuring sufficient clinical expertise to assess the accuracy and completeness of model-generated responses.
Details on each annotator's clinical training stage, medical specialty, and institutional affiliation at the time of annotation are provided in Supplementary Table~\ref{tab:annotator_profiles}.
We ultimately collected over 80,502 annotations across all stages of the pipeline, representing the largest human-labeled dataset for fine-grained evaluation of medical RAG systems.

\paragraph{Evidence Retrieval}
This subtask evaluates the relevance of retrieved passages with respect to the given query, defined by whether a passage contains information essential to support the answer.
Annotators are presented with a query, the top-16 passages retrieved by the system, and a set of must-have statements identified from the gold response. 
Each passage is paired with each must-have statement to form individual annotation units. 
For every passage-statement pair, annotators assess whether the passage provides sufficient information to support the given statement. 
\textit{Support} is defined as follows.
First, a statement is considered fully supported if the passage contains all the necessary information to infer the statement directly and unambiguously.
Second, partial support is assigned when the passage includes content that is thematically or contextually related to the statement, but additional assumptions or reliance on external knowledge are required to complete the inference.
Finally, a passage is labeled as providing no support if it fails to offer any relevant information for the statement, or if it contains content that contradicts it.
Illustrative examples for each support category are provided in Supplementary Table~\ref{tab:support_examples}. 
These examples were also included in the official annotation guidelines and used during annotator training to ensure consistency and shared understanding of the labeling criteria.

\paragraph{Evidence Selection}
This subtask evaluates the source attribution of model-generated references.
Annotators are given (1) a query, (2) the top-16 passages retrieved by the system, and (3) the references produced by the model.
For each reference, they determine which passage(s) served as its source of information.
A passage is considered a source if its metadata and content together provide sufficient information to reconstruct or justify the reference.
When multiple passages jointly contribute to a single reference, annotators are instructed to select all relevant passages.
If none of the retrieved passages offer adequate support, the reference is labeled as self-generated, indicating that the content originated from the model rather than from retrieved evidence.

\paragraph{Response Generation}

This subtask evaluates the quality of model-generated responses along two axes: factuality and completeness.
The factuality task assesses the accuracy of individual statements produced by the model. Annotators are provided with the patient query, the model-generated response, and the corresponding reference response for context. Each model statement is labeled as true if it aligns with the reference response or can reasonably be considered accurate based on the annotator's clinical expertise or authoritative medical sources. Conversely, a statement is labeled as false if it contradicts the reference, conflicts with established medical knowledge, or introduces implausible or unverifiable claims.
The completeness task examines whether the model response adequately captures the essential content of the reference response. Each reference response is decomposed into a set of must-have statements, i.e., statements deemed clinically important and expected to appear in a complete answer. 
For each must-have statement, annotators evaluate whether the model response fully supports it, partially supports it, or fails to support it. 
This task adopts the same three-way support classification scheme as the evidence retrieval task, enabling consistent, stage-aligned evaluation across the pipeline (see Supplementary Table~\ref{tab:support_examples} for examples).

\subsection{RAG Implementation}
\begin{figure}[t!]
\centering
\includegraphics[width=\linewidth]{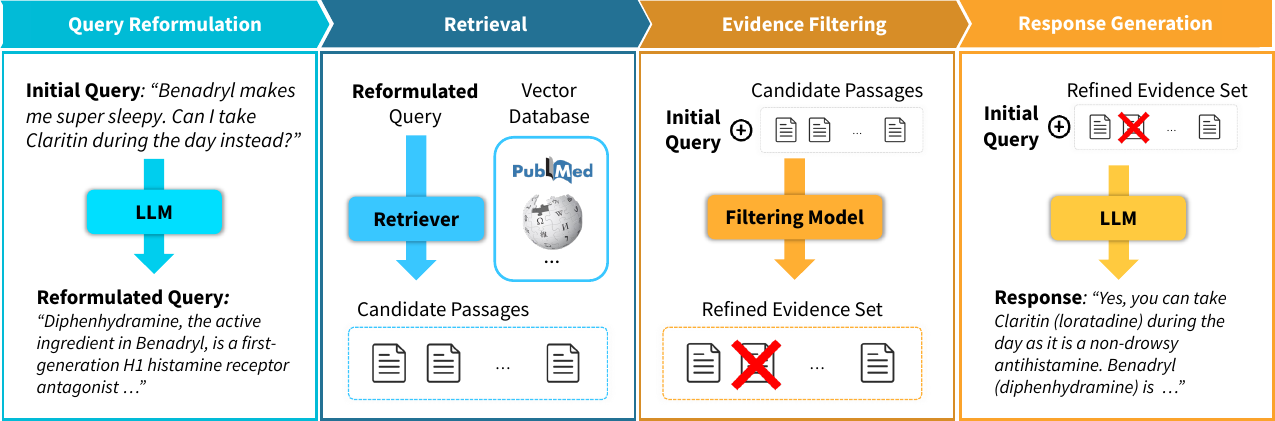}
\caption{
\textbf{Enhanced RAG pipeline.}
This introduces two lightweight components--query reformulation and evidence filtering--to improve the quality of retrieved evidence before generation. Given an input question, the LLM first reformulates it to better suit retrieval (Query Reformulation). The retriever then searches external knowledge sources such as PubMed or Wikipedia to gather candidate passages (Retrieval). A filtering model removes irrelevant evidence to retain informative evidence (Evidence Filtering). Finally, the LLM generates a response conditioned on the refined evidence set (Response Generation).
}
\label{figure:rag_pipeline}
\end{figure}

To design our RAG configuration, we conducted an extensive review of prior studies, particularly focusing on the choice of retrieval corpora and retriever (i.e., search engine) architectures. 
Rather than adopting a custom corpus or architecture, we opted for a widely adopted pipeline structure commonly reported in the literature.
A summary of this review is provided in Supplementary Table~\ref{tab:rag_summary}.
Building on this baseline RAG system, we incorporated two key components--query formulation and evidence filtering--to enhance performance. 
All implementation was based on the MedRAG toolkit (\url{https://github.com/Teddy-XiongGZ/MedRAG}), with additional functionalities newly developed for this study.
The resulting pipeline is illustrated in Fig.~\ref{figure:rag_pipeline}.

\paragraph{Retrieval Corpora}
We adopted the multi-source retrieval setup proposed by MedRAG~\cite{xiong2024benchmarking}, an early framework for medical-domain RAG, and further augmented it with clinical guidelines. 
Our final retrieval corpus comprises five complementary sources:
(1) PubMed: A widely used biomedical article repository containing 23.9 million articles, covering diverse topics across the biomedical and life sciences literature.
(2) StatPearls: A structured educational corpus comprising 301.2k passages from 9.3k peer-reviewed clinical articles, designed primarily for medical board exam preparation and continuing medical education. 
Each document follows a consistent format with sections such as epidemiology, pathophysiology, clinical presentation, and management, making it a valuable resource for evidence-grounded summarization and retrieval.
(3) Wikipedia: A broad-coverage knowledge source with 29.9 million passages from 6.5 million documents, covering both general-domain and medical topics.
(4) Medical textbooks: A set of 18 licensed medical textbooks spanning foundational medical disciplines, including anatomy, histology, neurology, pathology, physiology, biochemistry, immunology, obstetrics and gynecology, pediatrics, psychiatry, cell biology, internal medicine, pharmacology, and surgery. This corpus yielded 125.8k expert-authored snippets.
(5) Clinical guidelines: we additionally curated 723K passages from guidelines issued by nine leading health organizations and online clinical sources, including:
Cancer Care Ontario (CCO), Centers for Disease Control and Prevention (CDC), Canadian Medical Association (CMA), International Committee of the Red Cross (ICRC), National Institute for Health and Care Excellence (NICE), PubMed Clinical Queries, Strategy for Patient-Oriented Research (SPOR), World Health Organization (WHO), and WikiDoc.

\paragraph{Retriever}
We used MedCPT~\cite{jin2023medcpt}, a dual-encoder retriever pretrained on PubMed search logs. It comprises two components: a query encoder and an article encoder. 
The query encoder transforms an input query into a dense vector at inference time, while the article encoder pre-encodes all passages from the retrieval corpora into dense vector representations, which are stored in a vector database. 
During inference, the query vector is matched against the pre-encoded passage vectors using vector similarity metrics, and the top-k most similar passages are retrieved.

\paragraph{Evidence Filtering}
For the filtering model, we used Llama-3.1-8B to maintain architectural consistency with the base LLM. 
The model is designed to estimate $p(\hat{y}|q,d)$, where it takes a query $q$ and a candidate passage $d$ as input and predicts a binary label $\hat{y} \in \{0,1\}$, indicating whether the passage is relevant to the query.
We constructed a dataset of 3,200 query–passage pairs with expert-provided annotations serving as ground truth labels.
Given a query and a candidate passage, the model was prompted to generate one of two tokens: ``Yes'' if the passage contained supporting evidence for the query, or ``No'' otherwise (see Supplementary Table~\ref{tab:prompt_evidence_filtering} for the detailed prompt). 
The model was trained using standard causal language modeling loss (cross-entropy loss) over the next token.
Model hyperparameters were tuned through five-fold cross-validation.
The final configuration used a learning rate of 2e-6, a batch size of 8, and 3 training epochs.
Training was performed using the LLaMA-Factory repository~\cite{zheng2024llamafactory} on a single 80GB A100 GPU.

\paragraph{Query Reformulation}
In medical QA, the formulation of user queries plays a critical role in retrieval performance.
Overly detailed queries, such as those containing lengthy patient narratives or exhaustive symptom descriptions, may overwhelm the retriever with peripheral information, obscuring the core diagnostic clues.
Conversely, overly concise queries often lack sufficient clinical context, making it difficult to identify relevant evidence.
To address these challenges, we generate rationale queries that reflect the model's internal reasoning process for solving the clinical task.
This rationale serves as an optimized query representation:
for verbose inputs, it helps filter out irrelevant details and highlights diagnostically salient information;
for underspecified inputs, it fills in missing but medically necessary context that would be expected in a clinician’s reasoning.
Given an input question $q$, an LLM $M$ is prompted to produce a rationale $r$ as follows:
\begin{tcolorbox}
Respond to the following clinical decision-making task using the provided patient information, in a step-by-step fashion. Output your explanation and single option from the given options as the final answer.
\end{tcolorbox}
During retrieval, only the rationale $r$ is provided to the retriever.
Preliminary analysis showed that including both the original question and the rationale often exceeds the retriever's input length limit and may introduce redundant or conflicting signals.
We retrieve the top-k passages using $r$, and the same model $M$ is then used to generate the final response.

\subsection{Held-out Test Sets}

We used five medical-domain QA benchmarks that consist of exam-style multiple-choice questions and are widely used as standard testbeds in this field.
These datasets primarily provide answer keys without long-form explanations, and our evaluation focused solely on answer matching (i.e., accuracy). 
(1) MedQA~\cite{jin2021disease}:
A dataset of four-option multiple-choice questions collected from medical question banks and exam preparation sites.
(2) MedMCQA~\cite{pal2022medmcqa}:
A large-scale dataset of Indian postgraduate medical entrance exam questions (AIIMS, NEET-PG), covering a wide range of medical subjects.
(3) MMLU~\cite{hendrycks2021measuring}:
A popular, comprehensive multi-domain multiple-choice benchmark comprising 57 subjects.
We selected a subset of MMLU focused on clinical knowledge, college medicine, and professional medicine.
(4) MMLU-Pro~\cite{wang2024mmlu}:
An enhanced and more challenging version of MMLU that increases the number of answer choices per question (from 4 to 10), filters out trivial or noisy items, and emphasizes reasoning complexity. 
We likewise selected subsets of MMLU-Pro concentrated on clinical knowledge, college medicine, and professional medicine.
(5) MedXpertQA~\cite{zuo2025medxpertqa}:
A highly challenging medical QA benchmark comprising multiple-choice questions across 17 specialties and 11 body systems. It incorporates specialty board–style questions with extensive filtering, augmentation, and expert review. 
Even experts achieve only around 42\% accuracy, making it one of the most difficult medical QA benchmarks to date.
Due to inference cost constraints, we randomly sampled up to 500 examples per dataset when the original size exceeded this threshold.

\bibliography{references}

\section*{Acknowledgements}

This study is supported by the National Institutes of Health National Library of Medicine under Award Number R01LM014604.

\section*{Author Contributions Statement}

H.K. and Q.C. contributed to study design.
H.K., H.J., S.P., Y.P., J.P., and S.C. contributed to drafting and refining the annotation guidelines.
H.K. and J.S. configured the annotation interface.
S.A., H.J., S.P., Y.P., J.P., S.C., B.A.H.C., T.H., and J.Y.  contributed to data annotation (evidence retrieval and evidence selection). 
A.G., N.C., R.J., L.C., E.L., A.D., T.G., M.B.S., and Y.K. contributed to data annotation (factuality and completeness).
A.G., N.C., S.A., and R.J. contributed to qualitative analysis and case study. 
H.K., J.S., and E.F.W. contributed to model development. 
H.K. and Q.C. drafted the manuscript. 
R.A.A., J.Z., A.T., A.C., and H.X. provided critical feedback on the manuscript and study design.
All authors read and approved the final version of the manuscript.

\section*{Data/Code Availability}
Models and code are available at \url{https://github.com/Yale-BIDS-Chen-Lab/medical-rag}, and the full set of expert annotations will be released upon publication acceptance.

\section*{Competing Interests}
The authors declare no competing interests.

\appendix
\clearpage
\renewcommand{\tablename}{Supplementary Table}
\renewcommand{\thetable}{\arabic{table}}
\setcounter{table}{0}

\renewcommand{\figurename}{Supplementary Figure}
\renewcommand{\thefigure}{\arabic{figure}}
\setcounter{figure}{0}

\section*{Supplementary Information}

\AtBeginEnvironment{longtable}{\footnotesize}

\renewcommand{\arraystretch}{1.1}
\setlength{\tabcolsep}{4pt}

\begin{longtable}{p{1.3cm} p{2.3cm} p{3cm} p{2.2cm} p{3.6cm} p{3.5cm}}
\caption{\textbf{Summary of existing medical RAG frameworks.}
Frameworks are organized chronologically by publication date, summarizing their retriever and LLM configurations, underlying knowledge sources, and downstream medical applications. Only peer-reviewed studies published up to August 2025 were included, focusing specifically on medical and healthcare domains while excluding general biomedical NLP applications. 
Studies lacking sufficient implementation details were omitted. Each framework utilizes an LLM released after November 2022 (i.e., post-GPT-3.5). 
Descriptions are kept at a high level; although some frameworks draw from similar knowledge sources (e.g., clinical guidelines), their specific implementations and scopes vary across studies.
}
\label{tab:rag_summary} \\

\toprule
\textbf{Publication Date} & \textbf{Framework} & \textbf{Retriever(s)} & \textbf{LLM(s)} & \textbf{Knowledge Sources} & \textbf{Application(s)} \\
\midrule
\endfirsthead

\multicolumn{6}{c}{{\bfseries Table \thetable\ (continued)}} \\[3pt]
\toprule
\textbf{Publication Date} & \textbf{Framework} & \textbf{Retriever(s)} & \textbf{LLM(s)} & \textbf{Knowledge Sources} & \textbf{Application(s)} \\
\midrule
\endhead

\bottomrule
\endfoot

Jun 2023 & ChatDoctor~\cite{li2023chatdoctor} & Lexical matching & Llama & Wikipedia,  MedlinePlus & QA \\
Jul 2023 & accGPT~\cite{rau2023context} & Text-embedding-ada-002~\cite{openai2022vector} & GPT-3.5 & Clinical guidelines & Imaging recommendation \\
Aug 2023 & FraCChat~\cite{russe2023performance} & Text-embedding-ada-002 & GPT-3.5, GPT-4 & AO/OTA Fracture and Dislocation Classification Compendium (2018) & AO code identification \\
Dec 2023 & Clinfo.ai~\cite{lozano2023clinfo} & BM25, Entrez API~\cite{ncbi_entrez_2023} & GPT-3.5, GPT-4 & PubMed & QA, Summarization \\
Jan 2024 & Almanac~\cite{zakka2024almanac} & Text-embedding-ada-002   & GPT-4 & PubMed, Clinical guidelines, etc. & QA \\
Jan 2024 & RALL~\cite{guo2024retrieval} & DPR~\cite{karpukhin2020dense}, etc. & GPT-4, Llama-2 & Wikipedia, UMLS & Lay language generation \\
May 2024 & Ferber et al.~\cite{ferber2024gpt4guidelines} & Text-embedding-ada-002  & GPT-4   & Clinical guidelines  & QA \\
Jun 2024 & RECTIFIER~\cite{unlu2024retrieval} & Text-embedding-ada-002 & GPT-3.5, GPT-4 & Clinical guidelines & Clinical trial screening \\
Jul 2024 & Self-BioRAG~\cite{jeong2024improving} & MedCPT & Llama-2 & PubMed, PMC, Clinical guidelines, Textbooks & QA \\
Jul 2024 & DocOA~\cite{chen2024evaluating} & Not specified & GPT-3.5 & PubMed, Clinical guidelines & Osteoarthritis management \\
Aug 2024 & MedRAG~\cite{xiong2024benchmarking} & BM25, MedCPT, Contriever~\cite{izacard2021unsupervised}, SPECTER~\cite{cohan-etal-2020-specter} & GPT-3.5, GPT-4, Llama-2, Mixtral-8x7B, etc. & PubMed, StatPearls, Textbooks, Wikipedia & QA \\
Aug 2024 & Alkhalaf et al.~\cite{alkhalaf2024applyingrag} & BM25 & Llama-2 & EHR & Summarization\\ 
Sep 2024 & ChatZOC~\cite{luo2024development} & BM25 & Baichuan & Clinical guidelines, FAQs & Opthalmology QA \\
Sep 2024 & Benfenati et al.~\cite{benfenati2024nutrigeneticsrag} & BM25, GTE~\cite{li2023towards}& GPT-3.5, Mistral& Nutrigenetic polymorphism dataset, PubMed-derived nutrition–gene studies & Nutrigenetics QA\\
Oct 2024 & Bora and Cuayáhuitl~\cite{bora2024ragmedicalchatbot} & GTR-T5-Large & Llama-2, Mistral & Textbooks, Journals & QA \\
Dec 2024 & EyeGPT~\cite{chen_eyegpt_2024} & all-MiniLM-L6-v2~\cite{reimers-gurevych-2019-sentence} & Llama-2 & Textbooks, Custom Database & Ophthalmology QA \\
Jan 2025 & i-MedRAG~\cite{xiong2024improving} & MedCPT & GPT-4  & StatPearls, Textbooks & QA \\
Jan 2025 & Azimi et al.~\cite{bora2024ragmedicalchatbot} & Titan Text
Embeddings V2~\cite{aws2024titanTextEmbeddingsV2} & GPT-4o, Claude 3.5 Sonnet & Registered Dietitian (RD) exam questions across four nutrition domains (Academy of Nutrition and Dietetics guidelines and references) & Dietetic QA \\
Feb 2025 & Bailicai~\cite{long2024bailicaidomainoptimizedretrievalaugmentedgeneration} & MedCPT & Qwen~\cite{bai2023qwen}, Llama-2 & PubMed, StatPearls, Textbooks, Wikipedia & QA \\
Mar 2025 & Woo et al.~\cite{woo2025custom} & Internal retriever & GPT-3.5, GPT-4, Claude 3, Llama-3, Mistral-8×7B & Clinical guidelines & Clinical decision support \\
Apr 2025 & RAG squared~\cite{sohn2025rationale} & MedCPT & GPT-4o, Llama-3, etc. & PubMed, PMC, Clinical guidelines, Textbooks & QA \\
May 2025 & BriefContext~\cite{zhang2025leveraging} & BM25, MedCPT & GPT-4o, Llama-3.1 & PubMed, Textbooks & Summarization \\
May 2025 & MKRAG~\cite{shi2025mkrag} & Contriever, etc. & Vicuna-7B~\cite{chiang2023vicuna} & Disease Database & Knowledge graph QA \\
May 2025 & Yang et al.~\cite{yang_dual_2025} & Elasticsearch,  ColBERTv2~\cite{santhanam-etal-2022-colbertv2} & GPT-4, Claude 3 & Internal, expert-reviewed hospital documents & Clinical decision support \\
Jul 2025 & MedGraphRAG~\cite{wu-etal-2025-medical} & Graph-based retrieval & GPT-4, Gemini 1.0 Pro, Llama-2, Llama-3 & MIMIC-IV, FakeHealth, PubHealth, UMLS Graph & QA \\
Jul 2025 & RadioRAG~\cite{Tayebi_Arasteh_2025} & Text-embedding-ada-002 & GPT-4o mini & Radiopaedia & Case retrieval, Report generation \\
Jul 2025 & Wada et al.~\cite{wada_retrieval-augmented_2025} & Text-embedding-3-large\cite{openai2022vector}, Lexical matching & GPT-4o & Clinical guidelines (ACR Manual, ESUR Guidelines, etc.) & Safety consultation \\
Jul 2025 & MedOmniKB~\cite{chen-etal-2025-towards-omni} & MedCPT, Graph-based retrieval & GPT-4, Gemini 1.5 & PubMed, Clinical guidelines, Textbooks, Wikipedia, UMLS, DrugBank & QA \\
Aug 2025 & MedCoT-RAG~\cite{wang2025medcotragcausalchainofthoughtrag} & MedCPT & Llama-3.1 & PubMed, StatPearls, Textbooks, Wikipedia & QA \\
\end{longtable}

\clearpage

\newcommand{\cmark}{\textbf{\textcolor{teal}{\ding{51}}}} 
\newcommand{\xmark}{\textcolor{gray}{\ding{55}}} 

\begin{table}[ht!]
\centering
\footnotesize
\caption{\textbf{Accuracy of RAG configurations using different retrievers with Llama-3.1-8B, with or without evidence filtering and query reformulation.}
Each cell reports accuracy with 95\% confidence intervals. The non-RAG baseline is shown at the top.
Blue and red shading denote gains and drops from the non-RAG baseline, respectively.
}
\label{table:qa_result}
\begin{tabular}{lccccccc}
\toprule
 \multicolumn{3}{c}{\textbf{RAG Components}}  & \multicolumn{5}{c}{\textbf{Benchmarks}}  \\
\cmidrule(lr){1-3} \cmidrule(lr){4-8}
\textbf{Retrieval} & \begin{tabular}[c]{@{}c@{}}{\textbf{Filtering}}\\ \end{tabular}  & \begin{tabular}[c]{@{}c@{}}{\textbf{Query Reform.}}\\\end{tabular} & \textbf{MedQA} & \textbf{MedMCQA} & \textbf{MMLU} & \textbf{MMLU-Pro} & \textbf{MedXpertQA} \\
\midrule
\xmark~(Non-RAG) & \xmark & \xmark  & 0.672 & 0.528 & 0.738 & 0.620 & 0.136 \\
 & & & (0.630, 0.712) & (0.484, 0.572) & (0.700, 0.776) & (0.572, 0.668) & (0.106, 0.166) \\
\midrule
MedCPT & \xmark & \xmark & \cellcolor{decrease}
0.660 & \cellcolor{increase} 0.578 & \cellcolor{increase} 0.764 & \cellcolor{decrease} 0.583 & \cellcolor{increase} 0.179 \\
(k=16) & & & \cellcolor{decrease}(0.618, 0.702) & (\cellcolor{increase}0.534, 0.620) & \cellcolor{increase}(0.726, 0.800) & \cellcolor{decrease}(0.535, 0.634) & \cellcolor{increase}(0.146, 0.214) \\
 & \xmark & \cmark & \cellcolor{increase} 0.720 & \cellcolor{increase} \textbf{0.636} & \cellcolor{increase} 0.758 & \cellcolor{decrease} 0.583 & \cellcolor{increase} 0.172 \\
& & & \cellcolor{increase}(0.680, 0.758) & \cellcolor{increase}(0.594, 0.678) & \cellcolor{increase}(0.718, 0.796) & \cellcolor{decrease} (0.532, 0.634) & \cellcolor{increase}(0.140, 0.204) \\
 & \cmark & \xmark & \cellcolor{increase} 0.734 & \cellcolor{increase} 0.604 & \cellcolor{increase} \textbf{0.792} & \cellcolor{increase} \textbf{0.628} & \cellcolor{increase} 0.186 \\
& & & \cellcolor{increase} (0.696, 0.772) & \cellcolor{increase}(0.562, 0.648) & \cellcolor{increase}(0.756, 0.828) & \cellcolor{increase}(0.580, 0.679) & \cellcolor{increase}(0.152, 0.220) \\
 & \cmark & \cmark & \cellcolor{increase}\textbf{0.762} & \cellcolor{increase}0.622 & \cellcolor{increase}0.786 & \cellcolor{increase}\textbf{0.628} & \cellcolor{increase}\textbf{0.218} \\
& & & \cellcolor{increase}(0.724, 0.800) & \cellcolor{increase}(0.580, 0.664) & \cellcolor{increase}(0.750, 0.822) & \cellcolor{increase}(0.578, 0.676) & \cellcolor{increase}(0.182, 0.254) \\
\cmidrule(lr){2-8}
MedCPT & \xmark & \xmark & \cellcolor{decrease}0.636 & \cellcolor{increase}0.620 & \cellcolor{decrease}0.738 & \cellcolor{increase}\textbf{0.642} & \cellcolor{increase}0.162 \\
(k=32) & & & \cellcolor{decrease}(0.594, 0.678) & \cellcolor{increase}(0.578, 0.662) & \cellcolor{decrease}(0.698, 0.776) & \cellcolor{increase}(0.594, 0.690) & \cellcolor{increase}(0.130, 0.194) \\
 & \cmark & \xmark & \cellcolor{increase}0.702 & \cellcolor{increase}0.634 & \cellcolor{increase}0.760 & \cellcolor{decrease}0.602 & \cellcolor{increase}0.191 \\
& & & \cellcolor{increase}(0.660, 0.742) & \cellcolor{increase}(0.592, 0.676) & \cellcolor{increase}(0.722, 0.798) & \cellcolor{decrease}(0.551, 0.652) & \cellcolor{increase}(0.158, 0.226) \\
 & \xmark & \cmark & \cellcolor{increase}0.730 & \cellcolor{increase}0.630 & \cellcolor{increase}0.772 & \cellcolor{decrease}0.618 & \cellcolor{increase}0.186 \\
& & & \cellcolor{increase}(0.690, 0.768) & \cellcolor{increase}(0.586, 0.674) & \cellcolor{increase}(0.734, 0.808) & \cellcolor{decrease}(0.570, 0.666) & \cellcolor{increase}(0.152, 0.220) \\
 & \cmark & \cmark & \cellcolor{increase}\textbf{0.740} & \cellcolor{increase}\textbf{0.648} & \cellcolor{increase}\textbf{0.800} & \cellcolor{increase}0.631 & \cellcolor{increase}\textbf{0.206} \\
& & & \cellcolor{increase}(0.702, 0.778) & \cellcolor{increase}(0.606, 0.690) & \cellcolor{increase}(0.764, 0.834) & \cellcolor{increase}(0.583, 0.679) & \cellcolor{increase}(0.170, 0.242) \\
\midrule
BM25 & \xmark & \xmark & \cellcolor{decrease}0.610 & \cellcolor{increase}0.552 & \cellcolor{decrease}0.704 & \cellcolor{decrease}0.564 & \cellcolor{increase}0.150 \\
(k=16) & & & \cellcolor{decrease}(0.568, 0.654) & \cellcolor{increase}(0.508, 0.596) & \cellcolor{decrease}(0.662, 0.744) & \cellcolor{decrease}(0.513, 0.612) & \cellcolor{increase}(0.120, 0.182) \\
 & \cmark & \xmark & \cellcolor{decrease}0.642 & \cellcolor{increase}0.574 & \cellcolor{increase}0.748 & \cellcolor{increase}0.626 & \cellcolor{increase}\textbf{0.188} \\
& & & \cellcolor{decrease}(0.598, 0.684) & \cellcolor{increase}(0.530, 0.618) & \cellcolor{increase}(0.710, 0.786) & \cellcolor{increase}(0.578, 0.674) & \cellcolor{increase}(0.154, 0.222) \\
 & \xmark & \cmark & \cellcolor{increase}0.692 & \cellcolor{increase}0.560 & \cellcolor{increase}\textbf{0.765} & \cellcolor{increase}\textbf{0.687} & \cellcolor{increase}0.158 \\
& & & \cellcolor{increase}(0.652, 0.732) & \cellcolor{increase}(0.516, 0.604) & \cellcolor{increase}(0.728, 0.802) & \cellcolor{increase}(0.639, 0.733) & \cellcolor{increase}(0.126, 0.190) \\
 & \cmark & \cmark & \cellcolor{increase}\textbf{0.716} & \cellcolor{increase}\textbf{0.610} & \cellcolor{increase}0.744 & \cellcolor{increase}0.671 & \cellcolor{increase}0.166 \\
& & & \cellcolor{increase}(0.676, 0.754) & \cellcolor{increase}(0.568, 0.652) & \cellcolor{increase}(0.704, 0.782) & \cellcolor{increase}(0.623, 0.719) & \cellcolor{increase}(0.134, 0.200) \\
\cmidrule(lr){2-8}
BM25 & \xmark & \xmark & \cellcolor{decrease}0.648 & \cellcolor{increase}0.574 & \cellcolor{increase}0.749 & \cellcolor{decrease}0.593 & \cellcolor{increase}0.158 \\
(k=32) & & & \cellcolor{decrease}(0.606, 0.690) & \cellcolor{increase}(0.532, 0.618) & \cellcolor{increase}(0.712, 0.788) & \cellcolor{decrease}(0.543, 0.642) & \cellcolor{increase}(0.126, 0.190) \\
 & \cmark & \xmark & \cellcolor{increase}0.676 & \cellcolor{increase}0.584 & \cellcolor{increase}0.752 & \cellcolor{decrease}0.591 & \cellcolor{increase}\textbf{0.200} \\
& & & \cellcolor{increase}(0.634, 0.716) & \cellcolor{increase}(0.540, 0.628) & \cellcolor{increase}(0.712, 0.790) & \cellcolor{decrease}(0.543, 0.639) & \cellcolor{increase}(0.166, 0.236) \\
 & \xmark & \cmark &\cellcolor{increase} 0.718 & \cellcolor{increase}0.588 & \cellcolor{increase}\textbf{0.790} & \cellcolor{increase}0.653 & \cellcolor{increase}0.154 \\
& & & \cellcolor{increase}(0.678, 0.756) & \cellcolor{increase}(0.546, 0.630) & \cellcolor{increase}(0.752, 0.824) & \cellcolor{increase}(0.604, 0.703) & \cellcolor{increase}(0.122, 0.186) \\
 & \cmark & \cmark & \cellcolor{increase}\textbf{0.744} & \cellcolor{increase}\textbf{0.616} & \cellcolor{increase}0.780 & \cellcolor{increase}\textbf{0.663} & \cellcolor{increase}0.174 \\
& & & \cellcolor{increase}(0.706, 0.782) & \cellcolor{increase}(0.574, 0.658) & \cellcolor{increase}(0.742, 0.814) & \cellcolor{increase}(0.615, 0.711) & \cellcolor{increase}(0.142, 0.208) \\
\midrule
Qwen3-Emb-0.6B & \xmark & \xmark & \cellcolor{decrease}0.654 & \cellcolor{increase}0.596 & \cellcolor{decrease}0.738 & \cellcolor{decrease}0.612 & \cellcolor{increase}0.170 \\
(k=16) & & & \cellcolor{decrease}(0.612, 0.694) & \cellcolor{increase}(0.554, 0.638) & \cellcolor{decrease}(0.700, 0.776) & \cellcolor{decrease}(0.561, 0.660) & \cellcolor{increase}(0.138, 0.202) \\
 & \cmark & \xmark & \cellcolor{increase}0.698 & \cellcolor{increase}0.620 & \cellcolor{increase}0.764 & \cellcolor{increase}0.639 & \cellcolor{increase}0.182 \\
& & & \cellcolor{increase}(0.658, 0.736) & \cellcolor{increase}(0.576, 0.662) & \cellcolor{increase}(0.726, 0.802) & \cellcolor{increase}(0.588, 0.687) & \cellcolor{increase}(0.148, 0.216) \\
 & \xmark & \cmark & \cellcolor{increase}\textbf{0.748} & \cellcolor{increase}0.604 & \cellcolor{increase}0.760 & \cellcolor{increase}\textbf{0.668} & \cellcolor{increase}0.174 \\
& & & \cellcolor{increase}(0.710, 0.786) & \cellcolor{increase}(0.562, 0.646) & \cellcolor{increase}(0.722, 0.798) & \cellcolor{increase}(0.620, 0.717) & \cellcolor{increase}(0.142, 0.206) \\
 & \cmark & \cmark & \cellcolor{increase}0.734 & \cellcolor{increase}\textbf{0.630} & \cellcolor{increase}\textbf{0.768} & \cellcolor{increase}0.663 & \cellcolor{increase}\textbf{0.184} \\
& & & \cellcolor{increase}(0.694, 0.772) & \cellcolor{increase}(0.588, 0.672) & \cellcolor{increase}(0.730, 0.806) & \cellcolor{increase}(0.615, 0.711) & \cellcolor{increase}(0.150, 0.218) \\
\cmidrule(lr){2-8}
Qwen3-Emb-0.6B & \xmark & \xmark & \cellcolor{increase}0.696 & \cellcolor{increase}0.588 & \cellcolor{increase}0.746 & \cellcolor{decrease}0.602 & \cellcolor{increase}0.184 \\
(k=32) & & & \cellcolor{increase}(0.656, 0.736) & \cellcolor{increase}(0.544, 0.630) & \cellcolor{increase}(0.708, 0.784) & \cellcolor{decrease}(0.553, 0.652) & \cellcolor{increase}(0.150, 0.218) \\
 & \cmark & \xmark & \cellcolor{increase}0.702 & \cellcolor{increase}0.614 & \cellcolor{increase}0.758 & \cellcolor{increase}0.660 & \cellcolor{increase}0.178 \\
& & & \cellcolor{increase}(0.662, 0.742) & \cellcolor{increase}(0.572, 0.656) & \cellcolor{increase}(0.720, 0.796) & \cellcolor{increase}(0.612, 0.709) & \cellcolor{increase}(0.144, 0.212) \\
 & \xmark & \cmark & \cellcolor{increase}0.726 & \cellcolor{increase}0.600 & \cellcolor{increase}\textbf{0.764} & \cellcolor{increase}\textbf{0.676} & \cellcolor{increase}\textbf{0.196} \\
& & & \cellcolor{increase}(0.686, 0.764) & \cellcolor{increase}(0.556, 0.642) & \cellcolor{increase}(0.726, 0.802) & \cellcolor{increase}(0.628, 0.725) & \cellcolor{increase}(0.162, 0.230) \\
 & \cmark & \cmark & \cellcolor{increase}\textbf{0.760} & \cellcolor{increase}\textbf{0.630} & \cellcolor{increase}0.750 & \cellcolor{increase}0.655 & \cellcolor{increase}{0.192} \\
& & & \cellcolor{increase}(0.726, 0.796) & \cellcolor{increase}(0.586, 0.672) & \cellcolor{increase}(0.712, 0.788) & \cellcolor{increase}(0.607, 0.703) & \cellcolor{increase}(0.158, 0.226) \\
\bottomrule
\end{tabular}
\end{table}

\clearpage

\renewcommand{\arraystretch}{1.3}
\setlength{\tabcolsep}{6pt}

\begin{longtable}{p{0.95\linewidth}}
\caption{\textbf{Prompt for statement extraction.}
The \texttt{\{input\}} placeholder is intended to be replaced with the actual explanation or model response during prompting.
We used GPT-4o with temperature set to 0 for deterministic output.
}
\label{tab:prompt_stat_extraction}
\\
\hline
\textbf{Prompt Template} \\
\hline
\endfirsthead
\hline
\textbf{Prompt Template} \\
\hline
\endhead

Please breakdown the given text into independent facts. Review the examples provided below to gain a clearer understanding of the task requirements and the expected output format.
\\
Input: He made his acting debut in the film The Moon is the Sun’s Dream (1992), and continued to appear in small and supporting roles throughout the 1990s.
Output: [``He made his acting debut in the film.", ``He made his acting debut in The Moon is the Sun’s Dream.", ``The Moon is the Sun’s Dream is a film.", ``The Moon is the Sun’s Dream was released in 1992.", ``After his acting debut, he appeared in small and supporting roles.", ``After his acting debut, he appeared in small and supporting roles throughout the 1990s."]
\\
Input: He is also a successful producer and engineer, having worked with a wide variety of artists, including Willie Nelson, Tim McGraw, and Taylor Swift.
Output: [``He is successful.", ``He is a producer.", ``He is an engineer.", ``He has worked with a wide variety of artists.", ``illie Nelson is an artist.", ``He has worked with Willie Nelson.", ``Tim McGraw is an artist.", ``He has worked with Tim McGraw.", ``Taylor Swift is an artist.", ``He has worked with Taylor Swift."]
\\
Input: Possible causes for right lower abdominal pain in a young female are Appendicitis, Inflammatory bowel disease, Diverticulitis, Kidney stone, urinary tract infection, Ovarian cyst or torsion, Ectopic pregnancy, Pelvic inflammatory disease, endometriosis.
Output: [``Possible cause for right lower abdominal pain in a young female: Appendicitis.", ``Possible cause for right lower abdominal pain in a young female: Inflammatory bowel disease.", ``Possible cause for right lower abdominal pain in a young female: Diverticulitis.", ``Possible cause for right lower abdominal pain in a young female: Kidney stone.", ``Possible cause for right lower abdominal pain in a young female: urinary tract infection.", ``Possible cause for right lower abdominal pain in a young female: Ovarian cyst or torsion.", ``Possible cause for right lower abdominal pain in a young female: Ectopic pregnancy.", ``Possible cause for right lower abdominal pain in a young female: Pelvic inflammatory disease.",``Possible cause for right lower abdominal pain in a young female: endometriosis."]
\\
Input: Hep A IgM refers to a specific type of antibody called Immunoglobulin M (IgM) against the virus hepatitis A. When infected with hepatitis A, these antibodies are detectable at symptom onset and remain detectable for approximately three to six months. These antibodies might also be detectable in the first month after hepatitis A vaccination. A negative or non-reactive result means no IgM antibodies against hepatitis A found in your serum, meaning the absence of an acute or recent hepatitis A virus infection.
Output: [``Hep A IgM refers to a specific type of antibody called Immunoglobulin M (IgM) against the virus hepatitis A.", ``When infected with hepatitis A, these antibodies are detectable at the time of symptom onset.", ``When infected with hepatitis A, these antibodies remain detectable for approximately three to six months after infection.", ``These antibodies might also be detectable in the first month after hepatitis A vaccination.", ``The absence of IgM antibodies against hepatitis A in your serum indicates the absence of an acute or recent hepatitis A virus infection.", ``A negative or non-reactive result means that there were no IgM antibodies against hepatitis A found in your serum."]
\\
Input: methotrexate (Otrexup, Rasuvo, RediTrex) and thalidomide (Contergan, Thalomid) are both considered contraindicated for treatment of UC in pregnancy. possible treatment for UC during pregnancy include low-risk drugs such as aminosalicylates (sulfasalazine and mesalamine), immunomodulators (azathioprine, cyclosporine A ,6-mercaptopurine) and corticosteroids. Biological agents such as Infliximabl, Adalimumab, Vedolizumab and Ustekinumab is generally avoided during pregnancy as their safety in pregnancy is not well established yet.
Output: [``Methotrexate (Otrexup, Rasuvo, RediTrex) is contraindicated for treatment of ulcerative colitis in pregnancy.", ``Thalidomide (Contergan, Thalomid) is contraindicated for treatment of ulcerative colitis in pregnancy.", ``Aminosalicylates (sulfasalazine and mesalamine) are considered low-risk drugs for treatment of ulcerative colitis during pregnancy.", ``Immunomodulators (azathioprine, cyclosporine A, 6-mercaptopurine) are considered low-risk drugs for treatment of ulcerative colitis during pregnancy.", ``Corticosteroids are considered low-risk drugs for treatment of ulcerative colitis during pregnancy.", ``Treatment for ulcerative colitis during pregnancy with biological agents such as Adalimumab is generally avoided during pregnancy as their safety in pregnancy is not well established yet.", ``Treatment for ulcerative colitis during pregnancy with biological agents such as Vedolizumab is generally avoided during pregnancy as their safety in pregnancy is not well established yet.", ``Treatment for ulcerative colitis during pregnancy with biological agents such as Infliximab is generally avoided during pregnancy as their safety in pregnancy is not well established yet.", ``Treatment for ulcerative colitis during pregnancy with biological agents such as Ustekinumab is generally avoided during pregnancy as their safety in pregnancy is not well established yet."]
\\
\# YOUR TASK

Input: \texttt{\{input\}}

Output: 
\\
\hline
\end{longtable}

\clearpage
\renewcommand{\arraystretch}{1.3}
\setlength{\tabcolsep}{6pt}

\begin{longtable}{p{0.95\linewidth}}
\caption{\textbf{Prompt for must-have statement identification.}
The \texttt{\{query\}}, \texttt{\{answer\}}, and \texttt{\{statements\}} placeholders are intended to be replaced with the actual question, gold answer, and corresponding list of statements during prompting.
We used GPT-4o with temperature set to 0 for deterministic output.
}
\label{tab:prompt_must_have}
\\
\hline
\textbf{Prompt Template} \\
\hline
\endfirsthead
\hline
\textbf{Prompt Template} \\
\hline
\endhead

Please categorize the provided statements as either ``must-have" or ``nice-to-have".

*Must-have*: essential independent facts required to provide a complete answer to the given query.

*Nice-to-have*: supplementary or additional information that is useful but not essential.

Review the examples provided below to gain a clearer understanding of the task requirements and the expected output format.
\\
Query: I am a 33 years old female with right lower abdominal pain , what could it be?

Answer: Possible causes for right lower abdominal pain in a young female are Appendicitis, Inflammatory bowel disease, Diverticulitis, Kidney stone, urinary tract infection, Ovarian cyst or torsion, Ectopic pregnancy, Pelvic inflammatory disease, endometriosis.

Statement: [``Possible cause for right lower abdominal pain in a young female: Appendicitis.", ``Possible cause for right lower abdominal pain in a young female: Inflammatory bowel disease.", ``Possible cause for right lower abdominal pain in a young female: Diverticulitis.", ``Possible cause for right lower abdominal pain in a young female: Kidney stone.", ``Possible cause for right lower abdominal pain in a young female: urinary tract infection.", ``Possible cause for right lower abdominal pain in a young female: Ovarian cyst or torsion.", ``Possible cause for right lower abdominal pain in a young female: Ectopic pregnancy.", ``Possible cause for right lower abdominal pain in a young female: Pelvic inflammatory disease.", ``Possible cause for right lower abdominal pain in a young female: endometriosis."]

Output: [``Must-have", ``Must-have", ``Must-have", ``Must-have", ``Must-have", ``Must-have", ``Must-have", ``Must-have", ``Must-have"]
\\
Query: So what does the non reactive mean for the hep a igm

Answer: Hep A IgM refers to a specific type of antibody called Immunoglobulin M (IgM) against the virus hepatitis A. When infected with hepatitis A, these antibodies are detectable at symptom onset and remain detectable for approximately three to six months. These antibodies might also be detectable in the first month after hepatitis A vaccination. A negative or non-reactive result means no IgM antibodies against hepatitis A found in your serum, meaning the absence of an acute or recent hepatitis A virus infection.

Statements: [``Hep A IgM refers to a specific type of antibody called Immunoglobulin M (IgM) against the virus hepatitis A.", ``When infected with hepatitis A, these antibodies are detectable at the time of symptom onset.", ``When infected with hepatitis A, these antibodies remain detectable for approximately three to six months after infection.", ``These antibodies might also be detectable in the first month after hepatitis A vaccination.", ``The absence of IgM antibodies against hepatitis A in your serum indicates the absence of an acute or recent hepatitis A virus infection.", ``A negative or non-reactive result means that there were no IgM antibodies against hepatitis A found in your serum."]

Output: [``Must-have", ``Nice-to-have", ``Nice-to-have", ``Nice-to-have", ``Must-have", ``Must-have"]
\\
Query: What medications are contraindicated for a pregnant woman with ulcerative colitis?

Answer: methotrexate (Otrexup, Rasuvo, RediTrex) and thalidomide (Contergan, Thalomid) are both considered contraindicated for treatment of UC in pregnancy. possible treatment for UC during pregnancy include low-risk drugs such as aminosalicylates (sulfasalazine and mesalamine), immunomodulators (azathioprine, cyclosporine A ,6-mercaptopurine) and corticosteroids. Biological agents such as Infliximabl, Adalimumab, Vedolizumab and Ustekinumab is generally avoided during pregnancy as their safety in pregnancy is not well established yet. Treatment for ulcerative colitis during pregnancy should be tailored by your OBGYN and gastroenterologist.

Statements: [``Methotrexate (Otrexup, Rasuvo, RediTrex) is contraindicated for treatment of ulcerative colitis in pregnancy.", ``Thalidomide (Contergan, Thalomid) is contraindicated for treatment of ulcerative colitis in pregnancy.", ``Aminosalicylates (sulfasalazine and mesalamine) are considered low-risk drugs for treatment of ulcerative colitis during pregnancy.", ``Immunomodulators (azathioprine, cyclosporine A, 6-mercaptopurine) are considered low-risk drugs for treatment of ulcerative colitis during pregnancy.", ``Corticosteroids are considered low-risk drugs for treatment of ulcerative colitis during pregnancy.", ``Treatment for ulcerative colitis during pregnancy with biological agents such as Adalimumab is generally avoided during pregnancy as their safety in pregnancy is not well established yet.", ``Treatment for ulcerative colitis during pregnancy with biological agents such as Vedolizumab is generally avoided during pregnancy as their safety in pregnancy is not well established yet.", ``Treatment for ulcerative colitis during pregnancy with biological agents such as Infliximab is generally avoided during pregnancy as their safety in pregnancy is not well established yet.", ``Treatment for ulcerative colitis during pregnancy with biological agents such as Ustekinumab is generally avoided during pregnancy as their safety in pregnancy is not well established yet.", ``Treatment for ulcerative colitis during pregnancy should be tailored by your OBGYN and gastroenterologist."]

Output: [``Must-have", ``Must-have", ``Nice-to-have", ``Nice-to-have", ``Nice-to-have", ``Nice-to-have", ``Nice-to-have", ``Nice-to-have", ``Nice-to-have", ``Must-have"]
\\
\# YOUR TASK

Respond only with a list containing either `Must-have' or `Nice-to-have' for each statement. No other responses are required.

Query: \texttt{\{query\}}

Answer: \texttt{\{answer\}}

Statements: \texttt{\{statements\}}

Output: 
\\
\hline
\end{longtable}

\clearpage
\renewcommand{\arraystretch}{1.3}
\setlength{\tabcolsep}{6pt}

\begin{longtable}{p{0.12\linewidth} p{0.83\linewidth}}
\caption{\textbf{Prompt for response generation.}
Each prompt was constructed by combining a base query type (either patient queries or USMLE-style questions) with one or more template components shown in the table.
For instance, the complete prompt for the RAG configuration with patient queries consists of (Patient Query) + (RAG) + (Input Query).
}
\label{tab:prompt_response}
\\
\hline
\textbf{Prompt Type} & \textbf{Template} \\
\hline
\endfirsthead
\hline
\textbf{Prompt Type} & \textbf{Template} \\
\hline
\endhead
Patient Queries & Respond to the provided clinical inquiry. Your response must be accurate, clear, and include all essential information while omitting extraneous details. 
\\
\midrule
USMLE-style Queries & Answer to the provided multiple-choice question about medical knowledge in a step-by-step fashion. 
Output your explanation and single option from the given options as the final answer. 
\\
\midrule
Non-RAG & 
Additionally, distinguish between statements within your response that require authoritative references and those that do not. 
Here, a ``statement'' refers to an atomic or foundational piece of information, which may not necessarily form a complete sentence. For statements requiring references, include citations immediately after the respective statements, with reference numbers enclosed in square brackets (e.g., [1][2][3]). Citations may also be placed within the sentence, rather than just at the end, when appropriate. At the end of your response, provide a consolidated list of references, formatted according to AMA guidelines. Label the section as ``\#\#\# References,'' and number the references sequentially (1, 2, 3, etc.), with each reference on a separate line. Every reference listed MUST be appropriately cited within the response using square brackets. Please double-check thoroughly to ensure this requirement is met without exception. 
\\
\midrule
RAG & Additionally, distinguish between statements within your response that require authoritative references and those that do not. Here, a "statement" refers to an atomic or foundational piece of information, which may not necessarily form a complete sentence. For statements requiring references, include citations immediately after the respective statements, with reference numbers enclosed in square brackets (e.g., [1][2][3]). Citations may also be placed within the sentence, rather than just at the end, when appropriate.
\\
& The provided document snippets are retrieved through a search engine and may be used as references to support your response. External references not included in the provided materials may also be incorporated. At the end of your response, provide a consolidated list of references, formatted according to AMA guidelines. Label the section as ``\#\#\# References," and number the references sequentially (1, 2, 3, etc.), with each reference on a separate line. Every reference listed MUST be appropriately cited within the response using square brackets. Please double-check thoroughly to ensure this requirement is met without exception. 
\\
& - Document -

\texttt{\{passage\_1\}}

Metadata: \texttt{\{metadata\_1\}}
\\
& - Document -

\texttt{\{passage\_2\}}

Metadata: \texttt{\{metadata\_2\}}
\\
& ...
\\
& - Document -

\texttt{\{passage\_16\}}

Metadata: \texttt{\{metadata\_16\}}
\\
\midrule
Input Query & \#\#\# Input Query

\texttt{\{query\}}
\\
\hline
\end{longtable}

\clearpage
\renewcommand{\arraystretch}{1.3}
\setlength{\tabcolsep}{6pt}

\begin{longtable}{p{0.95\linewidth}}
\caption{\textbf{Prompt for model statement filtering.}
The \texttt{\{question\}} and \texttt{\{sentence\}} placeholders are intended to be replaced with the actual question or model statement during prompting.
We used GPT-4o with temperature set to 0 for deterministic output.
}
\label{tab:prompt_model_stat_filtering}
\\
\hline
\textbf{Prompt Template} \\
\hline
\endfirsthead
\hline
\textbf{Prompt Template} \\
\hline
\endhead

You are given a single sentence. Your task is to decide whether this sentence is distinctive or non-distinctive.

A sentence is Distinctive only if:

It does not simply restate or paraphrase the question stem, and

It introduces either:

(a) clinical reasoning or inference (e.g., "this pattern suggests a mechanical cause", "these findings are consistent with TSC"), or

(b) definitions or factual information that go beyond what is stated in the question, or 

(c) a final judgment or answer sentence (e.g., "the most accurate test is ~", "the answer is ~")
\\
A sentence is Non-Distinctive if it falls into any of the following:

It restates or rewords content already found in the question

It presents a raw observation or finding from the question (e.g., age, vitals, exam result)

It provides a definition or fact that is already included or implied in the question

It is procedural, meta, or instructional (e.g., “Let’s analyze…” or “We need to consider…”)
\\
You may refer to the list below when deciding. All of these are non-distinctive sentences:
\\
- To determine the most strongly associated condition for the patient described, we need to analyze the clinical information provided step-by-step.

- To determine the correct answer, let's analyze the given information step-by-step.

- To determine the most strongly associated condition with the patient's symptoms, we need to analyze the information provided.

- To address the query, we will analyze the patient's behavior and symptoms step-by-step to determine the most appropriate psychological defense mechanism being demonstrated.

- The patient's behavior will be analyzed to match the most appropriate psychological defense mechanism from the given options.

- To answer this question, we need to analyze the patient's behavior and determine which coping mechanism he is exhibiting.

- To approach this question, let's analyze the patient's behavior and the options provided.

- To determine the most accurate test for the condition described in the question, we need to consider the patient's symptoms.

- We need to consider physical examination findings.

- We need to consider laboratory test results in the context of common clinical scenarios.

- Step-by-Step Analysis: Patient Presentation.

- Step-by-Step Analysis: Laboratory Findings.
\\
Question: A 1-year-old girl is brought to a neurologist due to increasing seizure frequency over the past 2 months. She recently underwent a neurology evaluation which revealed hypsarrhythmia on electroencephalography (EEG) with a mix of slow waves, multifocal spikes, and asynchrony. Her parents have noticed the patient occasionally stiffens and spreads her arms at home. She was born at 38-weeks gestational age without complications. She has no other medical problems. Her medications consist of lamotrigine and valproic acid. Her temperature is 98.3°F (36.8°C), blood pressure is 90/75 mmHg, pulse is 94/min, and respirations are 22/min. Physical exam reveals innumerable hypopigmented macules on the skin and an irregularly shaped, thickened, and elevated plaque on the lower back. Which of the following is most strongly associated with this patient's condition? 

For the given question, all of these are non-distinctive sentences (not limited to the followings):

- The patient is a 1-year-old girl.

- The patient has a history of increasing seizure frequency.

- The EEG reveals hypsarrhythmia.

- The patient has numerous hypopigmented macules.

- The patient exhibits occasional stiffening and spreading of arms.

- The patient has an irregularly shaped, thickened, and elevated plaque on the lower back.

- The patient is currently on lamotrigine.

- The patient is currently on valproic acid. 

- The patient was born at term without complications.

- No other medical problems were reported for the patient.

- And more ...
\\
\# Question

\texttt{\{question\}}
\\
\# Target sentence (Do not output anything other than "Distinctive" or "Non-distinctive".)

\texttt{\{sentence\}}
\\
\hline
\end{longtable}

\clearpage
\renewcommand{\arraystretch}{1.3}
\setlength{\tabcolsep}{6pt}

\begin{longtable}{p{0.95\linewidth}}
\caption{\textbf{Prompt for statement-citation alignment.}
The \texttt{\{model\_response\}}, \texttt{\{model\_statements\}}, and \texttt{\{references\}} placeholders are intended to be replaced with the actual model response, corresponding list of statements, and references during prompting.
We used GPT-4o with temperature set to 0 for deterministic output.
}
\label{tab:prompt_alignment}
\\
\hline
\textbf{Prompt Template} \\
\hline
\endfirsthead
\hline
\textbf{Prompt Template} \\
\hline
\endhead

You are given a model-generated response that includes statements with inline citations (e.g., [1], [2], etc.), along with a list of references corresponding to those citations.
\\
Your task is to identify the alignment between each statement and the reference(s) it is associated with, based solely on the position of the citation markers in the response.
\\
Instructions:
\\
1. For each statement in the response, determine whether it is linked to one or more references by locating citation markers (e.g., [1]) near or within the statement.

2. For each linked reference, assign it to the corresponding statement.

3. Do not assess the factual accuracy or content of the reference. Only use the citation markers and their position in the response to make the alignment.

4. Consider a ``statement" to be a single sentence or a semantically independent clause.
\\
\#\#\# Input:

- **Model Response**:
\texttt{\{model\_response\}}

- **Statements**:
\texttt{\{model\_statements\}}

- **References**:
\texttt{\{references\}}
\\
\#\#\# Output format:
Your output should be organized in JSON, without any other responses, using the following format:
\\
\{
  ``\#1": \{
    ``statement": ``<copy of the statement>",
    ``refs": [1, 3] \/\/ or [] if no references are associated
  \},
  ``\#2": \{
    ``statement": ``...",
    ``refs": [...]
  \}
\}
\\
\#\#\# Output:
\\
\hline
\end{longtable}

\clearpage

\renewcommand{\arraystretch}{1.1}

\begin{longtable}{p{0.25\linewidth} p{0.4\linewidth} p{0.25\linewidth}}
\caption{\textbf{Profiles of annotators.} 
This table lists all clinicians and medical trainees who contributed to the expert annotation of evidence retrieval, selection, and response evaluation tasks, along with their affiliations and specialties. Affiliations are listed as of the project’s start and may have changed since then.}
\label{tab:annotator_profiles} \\
\toprule
\textbf{Name} & \textbf{Affiliation} & \textbf{Specialty (if applicable)} \\
\midrule
\endfirsthead

\multicolumn{3}{c}%
{{\bfseries Table \thetable\ (continued)}} \\[3pt]
\toprule
\textbf{Name} & \textbf{Affiliation} & \textbf{Specialty (if applicable)} \\
\midrule
\endhead

\endfoot

\multicolumn{3}{l}{\textbf{Residents / Fellows}} \\
Nicholas Cochran-Caggiano & Yale School of Medicine ({EMS Fellowship}) & Emergency Medicine \\
Leah Colucci & Yale School of Medicine & Emergency Medicine \\
Tuo Guo & Yale School of Medicine  & Emergency Medicine \\
Roy Jiang & Yale School of Medicine  & Dermatology \\
Eric Lai & Yale School of Medicine & Ophthalmology \\
Amisha Dave & Yale School of Medicine & Ophthalmology \\
Maxwell B. Singer & Yale School of Medicine & Ophthalmology \\
Aidan Gilson & Massachusetts Eye and Ear, Harvard Medical School & Ophthalmology \\
Yonghoe Koo & Asan Medical Center, University of Ulsan College of Medicine  & Physical Medicine and Rehabilitation \\
\addlinespace
\midrule
\multicolumn{3}{l}{\textbf{MD Candidates and Graduates}} \\
Serina Applebaum & Yale School of Medicine & – \\
Thomas Huang & Yale School of Medicine & – \\
Brittany Alexandra Herrera Contreras & Yale School of Medicine, San Juan Bautista School of Medicine & – \\
Heeju Jin & Seoul National University College of Medicine & – \\
Seihee Park & Seoul National University College of Medicine & – \\
Yujin Park & Seoul National University College of Medicine & – \\
Seoyoung Choi & Seoul National University College of Medicine & – \\
Jiyeong Park & Seoul National University College of Medicine & – \\
Jaehoon Yun & Hanyang University College of Medicine & – \\
\bottomrule
\end{longtable}

\clearpage

\renewcommand{\arraystretch}{1.3}
\setlength{\tabcolsep}{6pt}

\begin{longtable}{p{0.12\linewidth} p{0.83\linewidth}}
\caption{\textbf{Examples used in annotator training for support label decisions.}
These training samples illustrate how annotators were instructed to evaluate whether a given context supports the associated statement.}
\label{tab:support_examples}
\\
\hline
\textbf{Field} & \textbf{Content} \\
\hline
\endfirsthead
\hline
\textbf{Field} & \textbf{Content} \\
\hline
\endhead

\textbf{Context} & Individuals with diabetes are recommended to limit their consumption of sweets to one or two times per week. It is also suggested being selective with desserts and to focus on foods with a low glycemic index, such as high fiber foods like whole grains and legumes, as well as certain lower sugar fruits like berries, melons, and apples. \\
\cmidrule(lr){2-2}
\textbf{Statement} & It is recommended that diabetics avoid sweets. \\
\textbf{Explanation} & The context mentions to limit their consumption of sweets, which contrasts with the suggestion to avoid their consumption. \\
\textbf{Label} & \textbf{No Support} \\
\hline

\textbf{Context} & Right lower abdominal pain in a 25-year-old female could be caused by a variety of medical conditions. Some potential causes include: Ovarian cyst: a fluid-filled sac on the ovary—Ectopic pregnancy: a pregnancy that occurs outside the uterus. \\
\cmidrule(lr){2-2}
\textbf{Statement} & Possible cause for right lower abdominal pain in a young female can be Appendicitis. \\
\textbf{Explanation} & Insufficient information. \\
\textbf{Label} & \textbf{No Support} \\
\hline

\textbf{Context} & Mites and insects Ivermectin is also used to treat infection with parasitic arthropods. Scabies – infestation with the mite Sarcoptes scabiei – is most commonly treated with topical permethrin or oral ivermectin. For most scabies cases, ivermectin is used in a two dose regimen: a first dose kills the active mites, but not their eggs. Over the next week, the eggs hatch, and a second dose kills the newly hatched mites. For severe ``crusted scabies'', the U.S. Centers for Disease Control and Prevention (CDC) recommends up to seven doses of ivermectin over the course of a month, along with a topical antiparasitic. Both head lice and pubic lice can be treated with oral ivermectin, an ivermectin lotion applied directly to the affected area, or various other insecticides. Ivermectin is also used to treat rosacea and blepharitis, both of which can be caused or exacerbated by Demodex folliculorum mites. \\
\cmidrule(lr){2-2}
\textbf{Statement} & Permethrin cream (Elimite) is a topical treatment for scabies. \\
\textbf{Label} & \textbf{Full Support} \\
\hline

\textbf{Context} & Overall, the heart rate is decreased while stroke volume is increased, resulting in a net increase in blood pressure, leading to increased tissue perfusion. This causes the myocardium to work more efficiently, with optimized hemodynamics and an improved ventricular function curve. Other electrical effects include a brief initial increase in action potential, followed by a decrease as the K+ conductance increases due to increased intracellular amounts of Ca2+ ions. The refractory period of the atria and ventricles is decreased, while it increases in the sinoatrial and AV nodes. A less negative resting membrane potential is made, leading to increased irritability. The conduction velocity increases in the atria, but decreases in the AV node. The effect upon Purkinje fibers and ventricles is negligible. Automaticity is also increased in the atria, AV node, Purkinje fibers, and ventricles. \\
\cmidrule(lr){2-2}
\textbf{Statement} & A calcium channel blocker would not change the relative velocities of conduction in Purkinje fibers, atria, and ventricles. \\
\textbf{Explanation} & The context compares the pharmacological mechanism of action, action potential, and relative velocity after digoxin administration. However, it does not provide an explanation of conduction velocity when a calcium channel blocker is administered. While it is related to the statement in that it discusses conduction velocity after drug administration, it does not mention calcium channel blockers. Please be careful not to annotate it as partial support incorrectly. \\
\textbf{Label} & \textbf{No Support} \\
\hline

\textbf{Context} & Junctional rhythm describes an abnormal heart rhythm resulting from impulses coming from a locus of tissue in the area of the atrioventricular node, the ``junction'' between atria and ventricles. Under normal conditions, the heart's sinoatrial node determines the rate by which the organ beats – in other words, it is the heart's ``pacemaker''. The electrical activity of sinus rhythm originates in the sinoatrial node and depolarizes the atria. Current then passes from the atria through the atrioventricular node and into the bundle of His, from which it travels along Purkinje fibers to reach and depolarize the ventricles. This sinus rhythm is important because it ensures that the heart's atria reliably contract before the ventricles. \\
\textbf{Statement} & Conduction through the AV node is slow to allow the ventricles enough time to fill with blood. \\
\textbf{Explanation} & Insufficient information. \\
\textbf{Label} & \textbf{No Support} \\
\hline

\textbf{Context} & The Purkinje fibers (; Purkinje tissue or subendocardial branches) are located in the inner ventricular walls of the heart, just beneath the endocardium in a space called the subendocardium. The Purkinje fibers are specialized conducting fibers composed of electrically excitable cells. They are larger than cardiomyocytes with fewer myofibrils and many mitochondria. They conduct cardiac action potentials more quickly and efficiently than any other cells in the heart. Purkinje fibers allow the heart's conduction system to create synchronized contractions of its ventricles, and are essential for maintaining a consistent heart rhythm. Histology Purkinje fibers are a unique cardiac end-organ. Further histologic examination reveals that these fibers are split in ventricles walls. The electrical origin of atrial Purkinje fibers arrives from the sinoatrial node. Given no aberrant channels, the Purkinje fibers are distinctly shielded from each other by collagen or the cardiac skeleton. \\
\cmidrule(lr){2-2}
\textbf{Statement 1} & Conduction through the Purkinje system is the fastest within the heart. \\
\textbf{Label} & \textbf{Full Support} \\
\cmidrule(lr){2-2}
\textbf{Statement 2} & The conduction velocity through the heart in order of speed is Purkinje fibers > atria > ventricles > AV node. \\
\textbf{Explanation} & From the statement, ``\textit{They conduct cardiac action potentials more quickly and efficiently than any other cells in the heart},'' it is partially inferable that Purkinje fibers > others in terms of conduction speed and efficiency. \\
\textbf{Label} & \textbf{Partial Support} \\
\hline

\textbf{Context} & Function The AV node receives two inputs from the right atrium: posteriorly, via the crista terminalis, and anteriorly, via the interatrial septum. Contraction of heart muscle cells requires depolarization and repolarization of their cell membranes. Movement of ions across cell membranes causes these events. The cardiac conduction system (and AV node part of it) coordinates myocyte mechanical activity. A wave of excitation spreads out from the sinoatrial node through the atria along specialized conduction channels. This activates the AV node. The atrioventricular node delays impulses by approximately 0.09s. This delay in the cardiac pulse is extremely important: It ensures that the atria have ejected their blood into the ventricles first before the ventricles contract. This also protects the ventricles from excessively fast rate response to atrial arrhythmias. \\
\cmidrule(lr){2-2}
\textbf{Statement} & The conduction velocity of the structures of the heart are in the following order: Purkinje fibers > atria > ventricles > AV node. \\
\textbf{Explanation} & Insufficient information. Although there is some mention of ``delay,'' the context does not provide information to compare the conduction speeds of the AV node, Purkinje fibers, atria, and ventricles. Therefore, it should be annotated as no support. \\
\textbf{Label} & \textbf{No Support} \\
\hline

\textbf{Context} & Lexapro, also known by its generic name escitalopram, is a selective serotonin reuptake inhibitor (SSRI) commonly prescribed for the treatment of major depressive disorder (MDD) and generalized anxiety disorder (GAD) [1]. It works by increasing the levels of serotonin, a neurotransmitter associated with mood regulation, in the brain [2]. 

Dosage: The typical starting dose for adults with MDD or GAD is 10 mg once daily, which may be increased to a maximum of 20 mg per day depending on the patient's response and tolerability [4]. 

Side Effects: Common side effects include nausea, insomnia, fatigue, dry mouth, increased sweating, and sexual dysfunction [5]. Serious side effects can include serotonin syndrome, which is characterized by symptoms such as agitation, hallucinations, and rapid heart rate [6]. 

Contraindications: Lexapro should not be used in patients who are taking monoamine oxidase inhibitors (MAOIs) or have a known hypersensitivity to escitalopram or citalopram [7]. 

Warnings and Precautions: 

Suicidality: Antidepressants, including Lexapro, may increase the risk of suicidal thoughts and behaviors in children, adolescents, and young adults [8]. 

Serotonin Syndrome: Risk increases when used in combination with other serotonergic drugs [6]. 

QT Prolongation: Lexapro can cause dose-dependent QT interval prolongation, which can lead to serious cardiac arrhythmias [9].  \\
\cmidrule(lr){2-2}
\textbf{Statement 1} & Escitalopram is sold under the brand names Lexapro and Cipralex. \\
\textbf{Explanation} & From the statement ``\textit{Lexapro, also known by its generic name escitalopram},'' it is clear that Lexapro is the brand name for escitalopram. However, since there is no mention of Cipralex, it should be annotated as partial support. \\
\textbf{Label} & \textbf{Partial Support} \\
\cmidrule(lr){2-2}
\textbf{Statement 2} & Side effects of Escitalopram include GI symptoms such as nausea, diarrhoea, constipation. \\
\textbf{Explanation} & The context lists various side effects, but since there is no mention of diarrhea or constipation, it should be annotated as partial support. 
\\
\textbf{Label} & \textbf{Partial Support} \\
\cmidrule(lr){2-2}
\textbf{Statement 3} & Side effects of Escitalopram include insomnia. \\
\textbf{Label} & \textbf{Full Support} \\
\cmidrule(lr){2-2}
\textbf{Statement 4} & The FDA had published a black box warning regarding Escitalopram. \\
\textbf{Explanation} & Since ``\textit{The FDA had published a black box warning regarding Escitalopram}'' is not mentioned, Partial Support would be a more appropriate annotation. \\
\textbf{Label} & \textbf{Partial Support} \\
\cmidrule(lr){2-2}
\textbf{Statement 5} & Lexapro is not approved for use in pediatric patients less than 12 years of age. \\
\textbf{Explanation} & The context states, ``\textit{Antidepressants, including Lexapro, may increase the risk of suicidal thoughts and behaviors in children.}'' However, this does not indicate that the drug is not approved for use. Therefore, the appropriate annotation is no support. \\
\textbf{Label} & \textbf{No Support} \\
\hline

\textbf{Context} & Oral ivermectin is an effective option for the treatment of scabies. It is often used in a two-dose regimen, with the first dose killing the active mites and the second dose, administered one week later, targeting the newly hatched mites [1]. Ivermectin is particularly useful for treating crusted scabies and is sometimes prescribed in combination with a topical agent [2]. Although it is not FDA-approved for scabies treatment in the United States, it is widely used off-label for this purpose [3]. Studies have shown that oral ivermectin is effective and generally well-tolerated in children and infants, although its use in children weighing less than 15 kg is off-label [4, 5]. \\
\cmidrule(lr){2-2}
\textbf{Statement} & Ivermectin (Stromectol) is not safe for children under 15kg. \\
\textbf{Explanation} & Off-label use cannot always be considered unsafe. Therefore, this statement is not necessarily supported. \\
\textbf{Label} & \textbf{No Support} \\
\hline
\end{longtable}

\renewcommand{\arraystretch}{1.3}
\setlength{\tabcolsep}{6pt}

\begin{longtable}{p{0.95\linewidth}}
\caption{\textbf{Prompt for evidence filtering.}
}
\label{tab:prompt_evidence_filtering}
\\
\hline
\textbf{Prompt Template} \\
\hline
\endfirsthead
\hline
\textbf{Prompt Template} \\
\hline
\endhead

Given a query and a text passage, determine whether the passage contains supporting evidence for the query. Supporting evidence means that the passage provides clear, relevant, and factual information that directly backs or justifies the answer to the query.
\\
Respond with one of the following labels:
\\
``Yes" if the passage contains supporting evidence for the query.
\\
``No" if the passage does not contain supporting evidence.
\\
You should respond with only the label (Yes or No) without any additional explanation.
\\
Question: \texttt{\{question\}}
\\
Passage: \texttt{\{passage\}}
\\
\hline
\end{longtable}

\end{document}